\documentclass[smallextended]{svjour3}       

\smartqed  
\usepackage{graphicx,subfig,float}
\usepackage{amssymb}
\usepackage{amsmath} 
\usepackage{caption}
\usepackage{hyperref}

\usepackage{tikz}
\usepackage{xcolor}

\usepackage{xspace}

\newcommand{\D}{\mathbb{D}}
\newcommand{\x}[1]{x^{(#1)}}
\newcommand{\y}[1]{y^{(#1)}}

\newcommand{\R}[1]{\mathbb{R}^#1}

\newcommand{\E}[1]{\mathbb{E}_{#1}}
\newcommand{\Loss}[1]{\mathcal{L}_{#1}}
\newcommand{\z}[1]{z^{(#1)}}

\DeclareMathOperator{\EX}{\mathbb{E}}

\tikzset{latent/.style={black, draw=black, fill=green!15, circle}}
\tikzset{inputs/.style={black, draw=black, fill=green!15, circle}}
\tikzset{stats/.style={black, draw=black, fill=green!15, minimum width=10mm, minimum height=8mm, rectangle}}
\tikzset{ops/.style={black, draw=black, fill=red!20, circle}}
\tikzset{sample/.style={black, draw=black, fill=gray!20, diamond}}

\tikzset{bottomup/.style={red, very thick, ->}}
\tikzset{topdown/.style={blue, very thick, ->}}

\usepackage{import}
\usetikzlibrary{quotes,arrows.meta}
\usetikzlibrary{positioning}

\def\edgecolor{rgb:blue,4;red,1;green,4;black,3}
\newcommand{\midarrow}{\tikz \draw[-Stealth,line width =0.8mm,draw=\edgecolor] (-0.3,0) -- ++(0.3,0);}

\usepackage{Ball}
\usepackage{Box}
\usepackage{RightBandedBox}

\usetikzlibrary{positioning}
\usetikzlibrary{shapes.geometric}
\usetikzlibrary{3d} 

\def\ConvColor{rgb:yellow,5;red,2.5;white,5}

\def\PoolColor{rgb:red,1;black,0.3}

\def\FcColor{rgb:blue,2;green,5;white,5}

\begin{document}

\title{A survey on Variational Autoencoders from a GreenAI perspective}

\author{Andrea Asperti        \and
        Davide Evangelista    \and 
        Elena Loli Piccolomini 
}

\institute{A. Asperti \at
           University of Bologna, Department of Informatics: Science and Engineering (DISI) \\
           \email{andrea.asperti@unibo.it}           
           \and
           D. Evangelista \at
           University of Bologna, Department of Mathematics\\
           \email{davide.evangelista5@unibo.it}           
           \and
           E. {Loli Piccolomini} \at
           University of Bologna, Department of Informatics: Science and Engineering (DISI) \\
           \email{elena.loli@unibo.it}           
}

\date{Received: date / Accepted: date}

\maketitle

\begin{abstract}
Variational AutoEncoders (VAEs) are powerful generative models that merge elements from statistics and information theory with the flexibility offered by deep neural networks to efficiently solve the generation problem for high dimensional data. 
The key insight of VAEs is to learn the latent distribution
of data in such a way that new meaningful samples can be generated from it. This approach led to tremendous research and variations in the architectural design of VAEs, nourishing 
the recent field of research known as unsupervised representation learning. In this article, we provide a comparative evaluation of 
some of the most successful, recent variations of VAEs. 
We particularly focus the analysis on the energetic efficiency of
the different models, in the spirit of the so called Green AI, aiming both to
reduce the carbon footprint and the financial cost of generative 
techniques. For each architecture we provide its mathematical formulation, the ideas underlying its design, a detailed model description, a running implementation and quantitative results.
\end{abstract}

\section{Introduction}
Data generation, that is the task of generating new realistic samples given
a set of training data, is a fascinating problem of AI, with many relevant 
applications in different areas, spanning from computer vision, to natural language processing and medicine. Due to the curse of dimensionality, the problem
was practically hopeless to solve, until Deep Neural Networks enabled
the scalability of the required techniques via learned approximators. 
In recent years, deep generative models have
gained a lot of attention in the deep learning community, not just for their amazing applications, but also for the 
fundamental insight they provide on the encoding mechanisms of Neural Networks, the extraction of deep features, and the latent representation of
data. 

In spite of the successful results, deep generative modeling remains one
of the most complex and expensive tasks in AI. Training a complex 
generative model typically requires a lot of time and computational resources.
To make a couple of examples, the hyper-realistic Generative Adversarial Network
for face generation in \cite{InvidiaGAN18} required training on 8 Tesla V100 GPUs 
for 4 days; the training of BERT \cite{BERT}, a well known generative model for NLP, 
takes about 96 
hours on 64 TPU2 chips.

As remarked in \cite{GreenAI}, this computational cost has huge implications, both from the ecological point of view, and for the increasing difficulties
for academics, students, and researchers, in particular those from emerging economies, to do competitive, state of the art research.  
As a good practice in Deep Learning, one should give detailed reports about the financial cost of training and running models, in such a way to promote
the investigation of increasingly efficient methods. 

In this article, we offer a comparative evaluation of some recent generative
models. To make the investigation more focused and exhaustive, we restricted the analysis to a single class of models: the so called Variational Autoencoders \cite{Kingma13,RezendeMW14} (VAEs). 

Variational Autoencoders are becoming increasingly popular
inside the scientific community \cite{VAEbiomed,variationsVAE,astrovader}, both due to their
strong probabilistic foundation, that will be recalled 
in Section~\ref{sec:background}, and the precious insight on the latent
representation of data. However, in spite of the remarkable achievements, the behaviour of Variational Autoencoders is still far from satisfactory; there is a number
of well known theoretical and practical challenges that still hinder this generative paradigm (see Section~\ref{sec:vanillaVAE-implementation}), and whose solution drove
the recent research on this topic.
We try to give an exhaustive presentation of most of the VAE variants 
in the literature, relating them to the implementation and theoretical issues they were meant to address.

Hence, we focus on a restricted subset of recent architectures 
that, in our opinion, deserve a deeper investigation, for their
paradigmatic nature, the elegance of the underlying theory, 
or some key architectural insight. The three categories of
models that we shall compare are the Two-stage model \cite{TwoStage}, the Regularized 
Autoencoder\footnote{Strictly speaking, this is not a Variational model, but it helps in understanding them.} \cite{RegularizedAE}, and some versions of
Hierarchical Autoencoders. In the latter class, we provide a detailed analysis of the recent Nouveau VAE \cite{NVAE}; however, its complexity trespasses our computing facilities, so we investigate a much simpler model, and an interesting variant exploiting Feature-wise Linear Modulation \cite{Film} at high scales.

One of the metrics used to compare these models is their energetic efficiency, in the spirit of the emerging paradigm known 
as Green AI \cite{GreenAI}, aiming to assess performance/efficiency trade-offs.  Specifically, for each architecture, we provide a precise mathematical formulation, a discussion of the
main ideas underlying their design, a detailed model description, 
a running implementation in TensorFlow 2 freely available on our GitHub repository \href{https://github.com/devangelista2/GreenVAE}{https://github.com/devangelista2/GreenVAE}, and quantitative results.

\subsection{Structure of the article}\label{sec:structure}
The article is meant to offer a self-contained introduction 
to the topic of Variational Autoencoders, just assuming a basic
knowledge of neural networks. In Section~\ref{sec:background}
we start with the theoretical background, discussing the strong
and appealing probabilistic foundation of this class 
of generative models. In Section~\ref{sec:vanillaVAE-implementation} we address the way 
theory is translated into a vanilla neural net implementation, 
and introduce the many issues arising from this operation: 
balancing problems in the loss function (Section~\ref{sec:balancing}), 
posterior collapse (Section~\ref{sec:collapse}), aggregate posterior vs.
prior mismatch (Section~\ref{sec:mismatch}), blurriness (Section~\ref{sec:blurriness})
and disentanglement (Section~\ref{sec:disentanglement}). 

In the next three Sections we give a detailed mathematical introduction to the three classes of models for which we provide a deeper investigation, namely the 
Two-Stage approach in Section~\ref{sec:two-stage}, the regularized VAE in Section~\ref{sec:rae} and hierarchical models in Section~\ref{sec:HVAE}. Section~\ref{sec:setting} is devoted to 
describe our experimental setting: we discuss the metrics used for the 
comparison, and provide a detailed description of the neural networks
architectures. In Section~\ref{sec:tables} we provide the results
of our experimentation, making a critical discussion. In the conclusive Section~\ref{sec:conclusions} we summarize the content of the article
and draw a few considerations on the future of this field, and the
challenges ahead.

\section{Theoretical Background}\label{sec:background}
In this Section, we give a formal, theoretical introduction to Variational
Autoencoders (VAEs), deriving the so called Evidence Lower Bound (ELBO) adopted as a learning objective for this class of models. \smallskip

To deal with the problem of generating realistic data points 
$x \in \R{d}$ given a dataset $\D = \{ \x{1}, \dots, \x{N} \}$, 
generative models usually make the assumption that there exists a ground-truth distribution $\mu_{GT}$ 
supported on a low-dimensional manifold 
$\chi \subseteq \R{d}$ with dimension $k < d$, absolutely continuous with respect to the Hausdorff measure on $\chi$ and with density $p_{gt}(x)$. With this assumption, one can rewrite
\begin{align}
    p_{gt}(x) = \int_{\R{k}} p_{gt}(x, z) dz = \int_{\R{k}} p_{gt}(x|z)p(z) dz = \E{p(z)} [p_{gt}(x|z)]
\end{align}
where $z \in \R{k}$ is the latent variable associated with $x$, distributed with a simple distribution $p(z)$ named {\em prior distribution}. \\

The idea behind generative models is that if we can learn a good approximation of $p_{gt}(x|z)$ from the data, then we can use that approximation to generate new samples with ancestral sampling, that is:

\begin{itemize}
    \item Sample $z \sim p(z)$.
    \item Generate $x \sim p_{gt}(x|z)$.
\end{itemize}

For this reason, it is common to define a parametric family of probability distributions $\mathcal{P}_\theta = \{ p_\theta(x|z) | \theta \in \R{s} \}$ with a neural network, and to find $\theta^*$ such that
\begin{align}\label{MLE}
    \theta^* = \arg\max_\theta \E{\D}[\log p_\theta(x)] = \arg\max_\theta \E{\D} \Bigl[ \log \int_{\R{k}} p_\theta(x|z) p(z) dz \Bigr]
\end{align}
i.e. the Maximum Likelihood Estimation (MLE). \\

Unfortunately, (\ref{MLE}) is usually computationally infeasible. For this reason, VAEs define another probability distribution $q_\phi(z|x)$ named {\em encoder distribution} which describes the relationship between a data point $x \in \chi$ and its latent variable $z \in \R{k}$ and optimizes $\phi$ and $\theta$ such that:
\begin{align}
    \theta^*, \phi^* = \arg\min \E{\D} [D_{KL}(q_\phi(z|x) || p_\theta(z|x))]
\end{align}
where $D_{KL}(q_\phi(z|x) || p_\theta(z|x)) = \E{q_\phi(z|x)}[\log q_\phi(z|x) - \log p_\theta(z|x) ]$ is the Kullback-Leibler divergence between $q_\phi(z|x)$ and $p_\theta(z|x)$. \\

But
\begin{align}
    \nonumber &D_{KL}(q_\phi(z|x) || p_\theta(z|x)) \\
    \nonumber &=  \E{q_\phi(z|x)}[\log q_\phi(z|x) - \log p_\theta(z|x) ] \\  
   & =
    \E{q_\phi(z|x)}[\log q_\phi(z|x) - \log p_\theta(x|z) - \log p_\theta(z) + \log p_\theta(x) ] \\ 
    \nonumber &= D_{KL}(q_\phi(z|x) || p(z)) - \E{q_{\phi}(z|x)} [ \log p_\theta(x|z) ] + \log p_\theta(x)
\end{align}
Thus
\begin{align}
  \nonumber \E{q_{\phi}(z|x)} [ \log p_\theta(x|z) ] -  D_{KL}(q_\phi(z|x) || p(z)) & = \log p_\theta(x) - D_{KL}(q_\phi(z|x) || p_\theta(z|x)) \\ & \leq \log p_\theta(x)
\end{align}
since $D_{KL}(q_\phi(z|x) || p_\theta(z|x)) \geq 0$,
which implies that the Left Hand Side  of the equation above is a lower bound for the loglikelihood of $p_\theta(x)$. For this reason, it is usually called {\em ELBO} (Evidence Lower BOund). 

Since ELBO is more tractable than MLE, it is used as the cost function for the training of neural network in order to optimize both $\theta$ and $\phi$:
\begin{align}\label{eq:ELBO}
    &\Loss{\theta, \phi} (x) := \E{q_{\phi}(z|x)} [ \log p_\theta(x|z) ] -  D_{KL}(q_\phi(z|x) || p(z)) \\
    &\Loss{\theta, \phi} := \E{\D}[ \Loss{\theta, \phi} (x) ] 
\end{align}

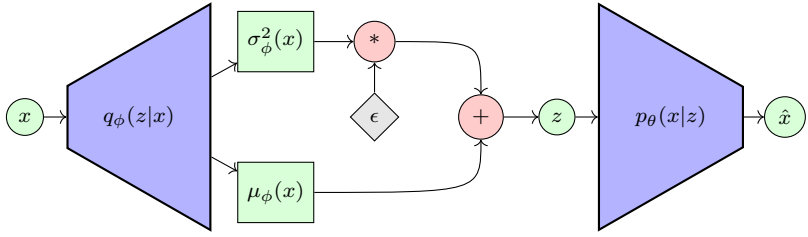
\begin{figure}
    \centering
    \begin{tikzpicture}
	\node[inputs] (X) at (0, 0) {$x$};
	\node[trapezium, trapezium angle=60, minimum width=30mm, fill=blue!30, draw, thick, shape border rotate=90] (Encoder) at (1.5, 0) {$q_\phi(z|x)$};
 	\node[stats] (sigma) at (3.3, 1) {$\sigma_\phi^2(x)$};
 	\node[stats] (mean) at (3.3, -1) {$\mu_\phi(x)$};
	
 	\node[ops] (times) at (4.6, 1) {*};
 	\node[ops] (add) at (6, 0) {+};
 	
 	\node[sample] (epsil) at (4.6, 0) {$\epsilon$};
 	
 	\node[inputs] (z) at (7, 0) {$z$};
	
 	\node[trapezium, trapezium angle=60, minimum width=30mm, fill=blue!30, draw, thick, shape border rotate=270] (Decoder) at (8.5, 0) {$p_\theta(x|z)$};
	
 	\node[inputs] (X_rec) at (10, 0) {$\hat{x}$};
 	
 	\draw[->] (X) -- (Encoder);
 	\draw[->] (Encoder) -- (sigma);
 	\draw[->] (Encoder) -- (mean);
    \draw[->] (sigma) -- (times);
    \draw[->] (times) .. controls (6, 1) .. (add);
    \draw[->] (mean) .. controls (6, -1) .. (add);
    \draw[->] (add) -- (z);
    \draw[->] (z) -- (Decoder);
    \draw[->] (Decoder) -- (X_rec);
    \draw[->] (epsil) -- (times);
    
\end{tikzpicture}
\label{fig:bottom_up}
    \caption{A diagram representing the VAE architecture. The stochastic component $\epsilon$ in the gray diamond is sampled from G(0,I). }
    \label{fig:vae}
\end{figure}

It is worth to remark that ELBO has a form resembling an autoencoder, where the term $q_\phi(z|x)$ maps the input $x$ to its latent representation $z$, and $p_\theta(x|z)$ decodes $z$ back to $x$. 
Figure \ref{fig:vae} shows a diagram representing the basic VAE structure.

For generative sampling, we forget the encoder and just exploit
the decoder, sampling the latent variables according to the prior 
distribution $p(z)$ (that must be known). 

\section{The vanilla VAE and its problems}\label{sec:vanillaVAE-implementation}
In this section, we explain how the theoretical form of the ELBO (eq. \ref{eq:ELBO}) can be translated into a numerical loss function 
exploitable for training of neural networks. This will allow us to
point out some of the typical problems that affect this architecture
and whose solution drove the design of the variants discussed in the 
sequel. 

In the vanilla VAE, we assume $q_\phi(z|x)$ to be 
a Gaussian (spherical) distribution $G(\mu_\phi(x),\sigma^2_\phi(x))$, 
so that learning $q_\phi(z|x)$ amounts to learning its two first
moments.

Similarly, we assume $p_\theta(x|z)$ has a Gaussian distribution around a decoder function $\mu_\theta(z)$. 
The functions $\mu_\phi(x)$, $\sigma^2_\phi(x)$ and $\mu_\theta(z)$ are modelled 
by deep neural networks. We remark that knowing the variance of latent variables allows sampling during training. 

If the model approximating the decoder function $\mu_\theta(z)$ is sufficiently expressive (that is case, for deep neural networks), the shape of the prior distribution $p(z)$ does not really matter, and
for simplicity it is assumed to be a normal distribution $p(z) = G(0,I)$.
The term $D_{KL}(q_\phi(z|X)||p(z))$ is hence the KL-divergence between two Gaussian distributions $G(\mu_\phi(x),\sigma^2_\phi(x))$ and $G(0, I)$ and it can be 
computed in closed form as:
\begin{equation}\label{eq:closed-form}
\begin{array}{l}
    D_{KL}(G(\mu_\phi(x),\sigma_\phi(x)),G(0,I)) = \\
    \hspace{1cm}\frac{1}{2} \sum_{i=1}^k \mu_\phi(x)^2_i + \sigma^2_\phi(x)_i-log(\sigma^2_\phi(x)_i) -1
\end{array}
\end{equation}
where $k$ is the dimension of the latent space. The previous equation has an intuitive explanation, as a cost function. 
By minimizing $\mu_\phi(x)$, when $x$ is varying on the whole dataset, 
we are centering the latent space around the origin (i.e. the mean of
the prior). The other component is preventing the variance 
$\sigma^2_\phi(x)$ to drop to zero, implicitly forcing a better 
coverage of the latent space. 

Coming to the reconstruction loss $\EX_{q_\phi(z|x)} [ \log p_\theta(x|z)]$, under the Gaussian assumption, the logarithm of $p_\theta(x|z)$ is the quadratic distance between $x$ and its reconstruction $\mu_\theta(z)$; the variance of this
Gaussian distribution can be understood as a parameter balancing
the relative importance between reconstruction error and KL-divergence
\cite{tutorial-VAE}.

The problem of integrating sampling 
with backpropagation during training is solved by the well known reparametrization trick proposed in  \cite{Kingma13,RezendeMW14}, where the sample is performed using 
a standard distribution (outside of the backpropagation flow) and
this value is rescaled with $\mu_\phi(x)$ and $\sigma_\phi(x)$.

The basic model of the Vanilla VAE that we just outlined 
is unfortunately hindered
by several known theoretical and practical challenges. 
In the next Sections, we give a short list of important 
topics which have been investigated in the literature, along 
with a short discussion of the main works addressing them.

\subsection{The balancing issue}\label{sec:balancing}
The VAE loss function is the sum of two 
distinct components, with somehow contrasting effects

\begin{equation}\label{eq:ELBO_split}
    \Loss{\theta, \phi} (x) := \underbrace{\E{q_{\phi}(z|x)} [ \log p_\theta(x|z) ]}_{\mbox{log-likelihood}} - \gamma  \underbrace{D_{KL}(q_\phi(z|x) || p(z))}_{\mbox{KL-divergence}}
\end{equation}

The log-likelihood loss is just meant to improve the quality of reconstruction, while the Kullback-Leibler component is acting as
a regularizer, pushing the aggregate
inference distribution $q_\phi(z) = \E{\D} [q_\phi(z|x)]$ 
towards the desired prior $p(z)$.

Log-likelihood and KL-divergence are frequently balanced by a suitable parameter, allowing to tune their mutual relevance.
The parameter is called $\gamma$, in this con, 
and it is considered as a normalizing factor for the reconstruction loss.

Privileging log-likelihood will improve the quality of reconstruction, neglecting the shape of the latent 
space (with ominous effects on generation). Privileging KL-divergence typically results in a smoother and normalized latent space, and more disentangled features \cite{beta-vae17,understanding-beta-vae18}; this usually
comes at the cost of a more noisy encoding, finally resulting in more blurriness in generated images. \cite{brokenELBOW}. 

Discovering a good balance between these components is a crucial aspect for an effective training of VAEs.

Several techniques for the calibration of $\gamma$ have been investigated in the literature, comprising an annealed optimization schedule \cite{Bowman15} or a policy 
enforcing minimum KL contribution from subsets of latent units \cite{autoregressive}. 
These schemes typically require hand-tuning and, 
as observed in \cite{overpruning17}, they easily risk to 
interfere with the principled regularization scheme that 
is at the core of VAEs.

An alternative possibility, investigated in \cite{TwoStage}, consists in {\em learning} the correct value for the balancing parameter during training, that also allows
its automatic calibration along the training process. 

In \cite{balancing} it is observed that, considering the objective function used in \cite{TwoStage} in order to learn $\gamma$, 
the optimal $\gamma$ parameter is in fact proportional to the current reconstruction error; so learning can be replaced
by a mere computation, using e.g. a running average. This has a
simple and intuitive explanation: what matters is to try
to maintain a {\em fixed} balance between the two components
during training: if the reconstruction error decreases, 
we must proportionally decrease the KL component that could
otherwise prevail, preventing further improvements. 
The technique in \cite{balancing} is simple and effective: we
shall implicitly adopt it in all our VAE models, unless 
explicitly stated differently.

A similar technique has been recently investigated in 
\cite{controlvae}, where the KL-divergence is used as a feedback during model training for dynamically tuning
the balance of the two components.

\subsection{Variable collapse phenomenon}\label{sec:collapse}
The KL-divergence component of the VAE loss function typically induces a parsimonious use of latent variables, some of which
may be altogether neglected by the decoder, possibly resulting in an under-exploitation
of the network capacity; if this is a beneficial side effect or regularization \cite{sparsity,TwoStage}
or an issue to be solved (\cite{BurdaGS15,overpruning17,Trippe18,PosteriorCollapse}), it is still debated.

\begin{figure}[ht]
\includegraphics[width=.5\textwidth]{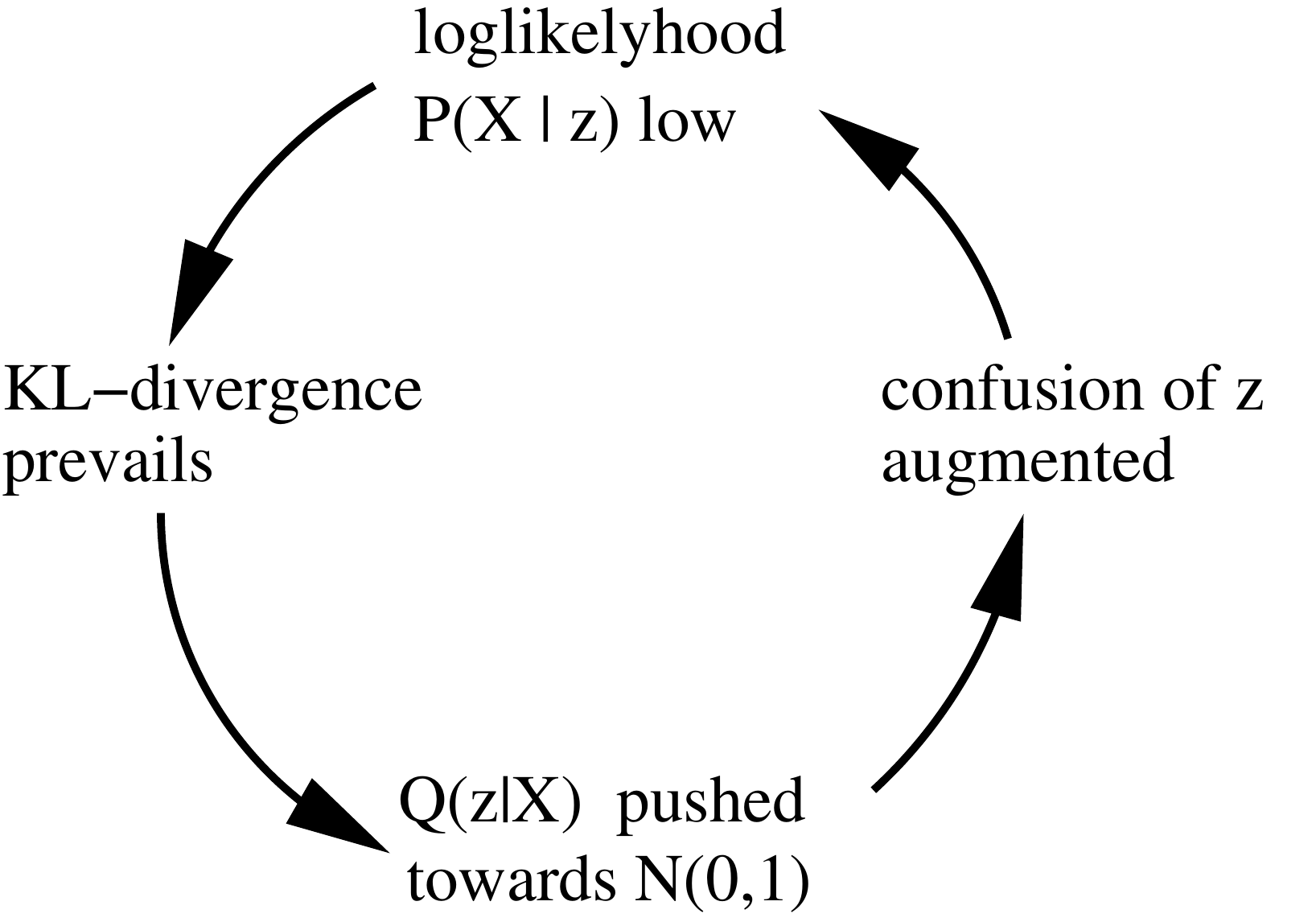}
\includegraphics[width=.5\textwidth]{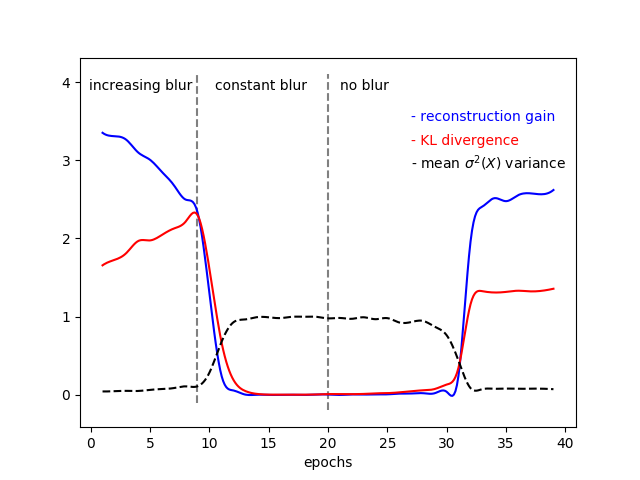} 
\caption{(Left) The vicious cycle leading to the variable collapse. 
(Right) An empirical demonstration of the phenomenon: we apply a progressive 
noise to a latent variable, reducing its contribution to reconstruction; at some
point, KL-divergence prevails, enlarging the sampling variance of the variable
and making it even more noisy; the phenomenon has catastrophic nature, leading to 
a complete collapse of the variable. If we remove the artificial noise, the variable gets reactivated. Pictures borrowed from \cite{collapse}.}
\label{fig:collapse}
\end{figure}

The variable collapse phenomenon has a quite intuitive explanation. If, during training, a latent variable gives a modest contribution for the reconstruction of the
input (in comparison with other variables), then the Kullback-Leibler divergence may prevail, pushing the mean towards 0 and the standard deviation towards 1.
This will make the latent variable even more noisy, in a vicious cycle that will eventually
induce the network to completely ignore the latent variable (see Figure~\ref{fig:collapse}, Left).

As described in \cite{collapse}, one can easily get an empirical evidence of the phenomenon by adding some artificial noise
to a variable and monitoring its evolution during
training (Figure~\ref{fig:collapse}, Right).
The contribution of a latent variable to reconstruction is computed as the difference between the reconstruction loss when the variable is masked with respect to the case when it is normally taken into account; we call this information {\em reconstruction gain}.

When the reconstruction gain of the variable is becoming less than the KL-divergence,
the variable gets ignored by the network: its correspondent mean value  will collapse to 0
(independently from $x$) and its  sampling variance is pushed to 1. 
Sampling has no impact on the network, precisely because the variable is ignored by the
decoder.

The variable collapse phenomenon is, at some extent, reversible. However, reactivating
a collapsed variable is not a completely trivial operation for a network, probably due
to saturation effects and vanishing gradients. 

\subsection{Aggregate posterior vs. expected prior mismatch}
\label{sec:mismatch}
The crucial point of VAEs is to learn an encoder  producing an 
{\em aggregate posterior distribution} $q_\phi(z)  = \E{\D} [q_\phi(z|x)]$ close to the prior $p(z)$. If this objective
is not achieved, generation is doomed to fail.

Before investigating ways to check the intended behavior, let us
discuss how the Kullback-Leibler divergence  term in \eqref{eq:ELBO_split} acts on the distance  $q_\phi(z)$ and $p(z)$.
So, let us average over all $x$ (we omit the $\phi$ subscript):
\begin{equation}\label{eq:averaging}
  \begin{array}{ll}
    \EX_{p_{gt}(x)} [D_{KL}(q(z|x)|p(z))] \smallskip\\\smallskip
    = - \EX_{p_{gt}(x)} [\mathcal{H}(q(z|x))] + \EX_{p_{gt}(x)} [\mathcal{H}(q(z|x),p(z))] &  \mbox{by def. of KL}\\\smallskip
    = - \EX_{p_{gt}(x)} [\mathcal{H}(q(z|x))] + \EX_{p_{gt}(x)} [\EX_{q(z|x)} [\log p(z)]] & \mbox{by def. of entropy}\\\smallskip
    = - \EX_{p_{gt}(x)} [\mathcal{H}(q(z|x))] + \EX_{q(z)} [\log p(z)] & \mbox{by marginalization}\\\smallskip
    = - \underbrace{\EX_{p_{gt}(x)}[ \mathcal{H}(q(z|x))]}_{\substack{\mbox{Avg. Entropy}\\\mbox{of } q(z|x)}} +
          \underbrace{\mathcal{H}(q(z),p(z))}_{\substack{\mbox{Cross-entropy of }\\q(x) \mbox{ vs }p(z)}} & \mbox{by def. of entropy}
  \end{array}
  \end{equation}

By minimizing the cross-entropy between $q(z)$ and $p(z)$
we are pushing one towards the other. Jointly, we try to augment the entropy of $q(z|x)$; under the assumption that $q(z|x)$ is Gaussian,
its entropy is $\frac{1}{2}log(e\pi\sigma^2)$: we are thus enlarging
the (mean) variance, further improving the coverage of the 
latent space, essential for generative sampling.

As a simple sanity check, one should always monitor the moments of 
the aggregate posterior distribution $q(z)$ during training: the
mean should be 0, and the variance 1. Since collapsed variables
could invalidate this computation (both mean and variance are close to 0), it is better 
to use an alternative rule \cite{aboutVAE} : if we look at 
$q(z) = \EX_{p_{gt}(x)} [q(z|x)]$ as a Gaussian Mixture Model (GMM), its variance $\sigma_{GMM}^2$ is given by the sum
of the variances of the means $\EX_{p_{gt}(x)} [\mu_{\phi}(x)^2]$ and the mean of the variances $\EX_{p_{gt}(x)} [\sigma_{\phi}^2(x)]$
of the components (supposing that $\EX_{p_{gt}(x)} [\mu_{\phi}(x)$]=0):
\begin{equation}
\sigma_{GMM}^2 = \EX_{p_{gt}(x)} [\mu_{\phi}(x)^2] +  \EX_{p_{gt}(x)} [\sigma_{\phi}^2(x)]  = 1
\end{equation}
where in this case $\mu_\phi(x)$ and $\sigma_\phi^2(x)$ are the values computed by the encoder. 

This is called {\em variance law} in \cite{aboutVAE}, and can be
used to verify that the regularization effect of the KL-divergence is properly working. 

The big problem is that, even if the two first moments of $q(z)$
are 0 and 1, this does not imply that it should look like a Normal (meaning
that the KL-divergence got lost in some local minimum, contenting 
itself with adjusting the first moments of the distributions).

The potential mismatch between $q(z)$ and the expected prior $p(z)$ is a problematic aspect of VAEs that, 
as observed by many authors \cite{ELBOsurgery,rosca2018distribution,aboutVAE}, could seriously compromise
the whole generative framework. Attempts to solve this issue have been
made both by acting on the loss function \cite{WAE} or by exploiting more complex priors \cite{autoregressive,Vamp,resampledPriors}. 

An interesting possibility, that has been recently 
deployed in the Hyperspherical VAE \cite{HypersphericalVAE}, consists in 
replacing the Gaussian Distribution with the von Mises-Fisher (vMF) distribution \cite{fisher}, that is a continuous distribution on the N-dimensional sphere in use in {\em directional statistics}.

An orthogonal, drastic alternative consists in renouncing to 
work in the comfortable setting of continuous latent variables, passing instead in the discrete domain. This approach is at the core of the Vector Quantized VAE \cite{VQ-VAE} (VQ-VAE): each latent variable is forced to occupy a position in a finitely sampled space, so that we can treat each latent variable as a $k$-dimensional vector in a space of dimension $d$. This discrete encoding is exploited
during sampling, where the prior is learnt via a suitable autoregressive technique.

\subsubsection{Clustering, GMM and Two-stage}
In case input data are divided in subcategories (as in the case of MNIST and Cifar10), or have macroscopic attributes
like, say, a different color for hairs in the case of 
CelebA, we could naturally expect to observe this information
in the latent encoding of data \cite{clustering}. 
In other words, we could 
imagine the latent space to be organized in 
{\em clusters}, (possibly) reflecting macroscopic features
of data. 

\begin{figure}[ht]
\begin{center}
\includegraphics[width=.6\textwidth,height=.45\textwidth]{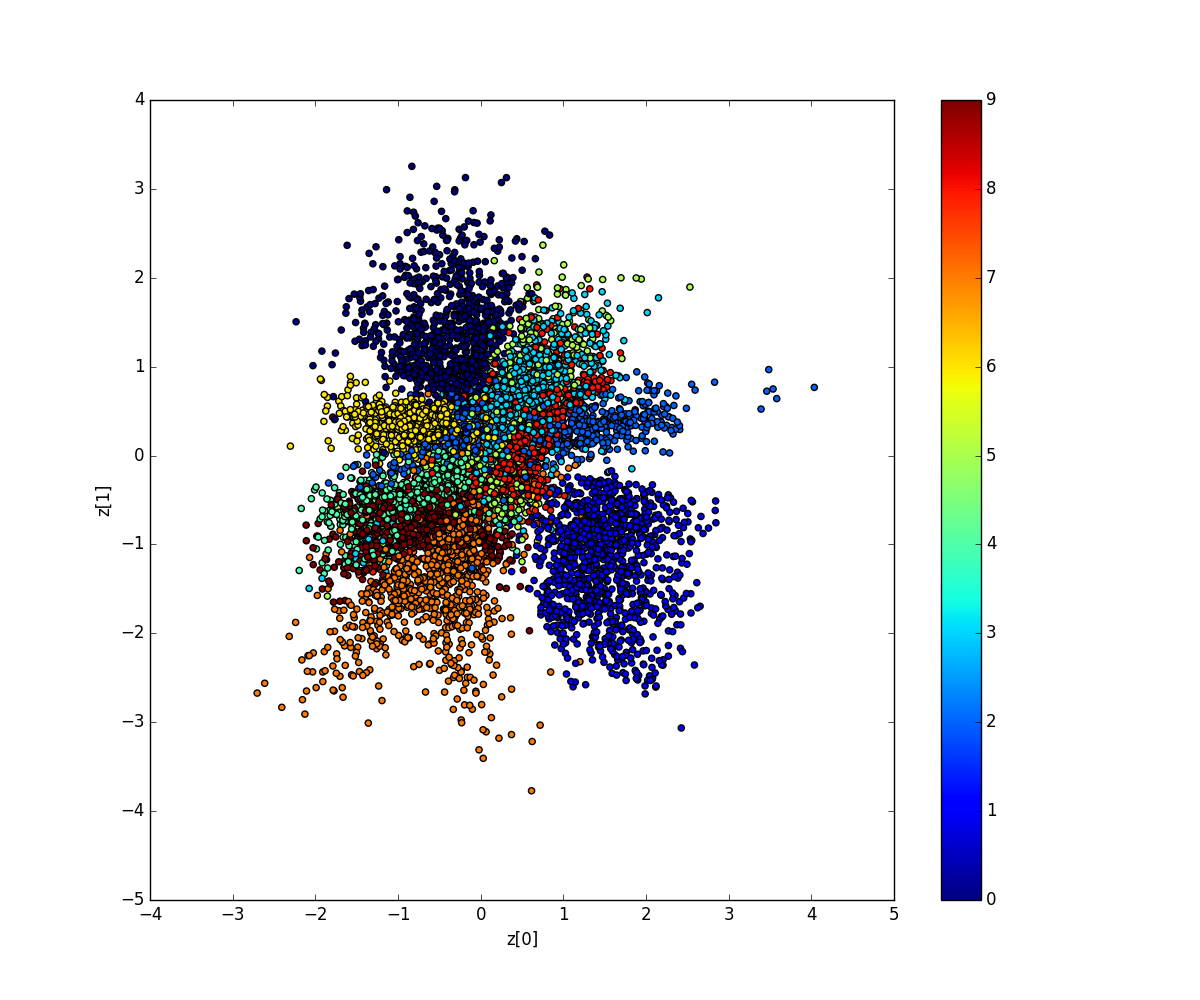}
\caption{\label{fig:mnist-latent}Latent encoding of MNIST digits in a latent space of dimension 2. Digits in different categories are represented with a different color. Observe: (1) the overall (rough) Gaussian-like disposition of all digits and (2) the typical organization in clusters, in contrast with the uni-modal objective of KL-regularization.}
\end{center}
\end{figure}

To make an example, in Figure \ref{fig:mnist-latent} it is described the latent encoding of MNIST digits, with a different color for each class in the range 0-9.

We can clearly observe that different digits naturally organize themselves in separate clusters. While the overall
distribution still has a Gaussian-like shape, the 
presence of clusters may obviously contrast with the required
smoothness of the internal encoding, introducing regions  with higher/lower probability densities. Observe e.g. the gaps between some of the clusters: sampling in such a region will eventually result in a poor generative output. In other words, clustering could be one of the main source for the mismatch between the prior and the aggregate posterior.

While the phenomenon is evident in a low-dimensional setting, 
it is more difficult to observe and testify it in higher dimensions. Remember that one of the VAE assumptions is that, as far as you have a sufficiently expressive decoder, the prior does not really
matter since the decoder will be able to turn each distribution into the desired one \cite{tutorial-VAE}.

Still, it makes sense to try to exploit clustering, and a 
natural approach consists in using a GMM model.
Several works have been done in this direction. The simplest 
approach, followed in \cite{RegularizedAE}, is to superimpose a
GMM of fixed dimension on the latent space via ex-post estimation 
using standard machine learning techniques (this is also the 
approach we shall follow in some of our tests). Alternatively, 
the GMM model can be {\em learned}. In the Variational 
Deep Embedding approach \cite{clustering} (VaDE), that essentially provides an unsupervised clustering model, the relevant
statistics of the GMM are estimated via Maximum Likelihood Estimation, in a way similar to the Vanilla case (see also
\cite{deepGMM} for a similar, slightly more sophisticated approach).

In the so called Two-Stage model \cite{TwoStage} a second VAE is trained to learn an accurate approximation of $q(z)$; 
samples from a Normal distribution are first used to generate samples
of $q(z)$, passed to the actual generator of data points. We shall give an extensive discussion of to the Two-Stage approach in 
Section \ref{sec:two-stage}.

In \cite{deterministic}, it is proposed to give an ex-post estimation of $q(z)$, e.g. imposing  a distribution with a sufficient complexity (they consider a combination of 10 Gaussians, reflecting the ten categories of MNIST and Cifar10). A suitable regularization technique alternative to KL is used to induce the desirable smoothness of the latent space. A deeper analysis of this approach is done in 
Section \ref{sec:rae}.

An additional and interesting issue of the Two-Stage model concerns the similarity measure to use as a loss function in the second stage. In \cite{TwoStage}, the traditional mean squared error and categorical
cross entropy are considered. However, we discovered that 
{\em cosine distance} works amazingly better. We did not get to 
cosine distance by trial and error, but by a long and deep 
investigation on latent representations. These results will be 
the object of a forthcoming article.

\subsection{Blurriness}\label{sec:blurriness}
Variational Autoencoders (VAEs), in comparison with alternative generative techniques, usually produce images with a characteristic and annoying blurriness. 
The phenomenon can also be observed in terms of the mean variance of pixels in generated images, which is significantly lower than that
for data in the training set \cite{varianceloss}.

The source of the problem is not easy to identify, but it is likely 
due to {\em averaging}, implicitly underlying the VAE frameworks 
(and, more generally, the whole autoencoder approach).
In presence of multimodal output, a loglikelihood objective typically results in averaging and hence blurriness \cite{tutorial-GAN}. 

Variational Autoencoders are intrinsically multimodal, both 
due to dimensionality reduction, and to the sampling process during
training.

Several attempts to solve the issue acting on the reconstruction metrics
have been made. Structural similarity (frequently used for deblurring purposes) does not seem to be effective \cite{DosovitskiyB16}. 
Better results can be obtained by considering deep hidden features
extracted from a pretrained image classification model, like e.g.
VGG19 \cite{DeepFeaturesVAE}.
In models of the VAE-GAN family \cite{vaegan,F-VAEGAN-D2,Zero-VAE-GAN}, the reconstruction loss is altogether replaced by a discriminator trying to distinguish real images from generated ones.
The use of a discriminator, assessing the quality of generated data
and acting on the density of the prior, 
is also a basic component of the recent VAEPP model (VAEs with a pullback
prior) \cite{pullbackPrior}. 
 
The most promising approaches are however based on iterative/hierarchical approaches \cite{DRAW,Eslami18,NVAE}.
In these architectures, following the idea of latent Gaussian models 
\cite{recurrent_latent},
the vector of latent variables $z$ is split
into $L$ groups of latent variables $z_l , l = 1,...,L$ and the density over the variable of interest is constructed sequentially, in terms of
latent variables of lower indices.
For instance, the prior $p(z)$ would be written as an autoregressive density of the following kind:
\begin{equation}
    p(z) = \prod_{l=1}^L p_l(z_l|z_{<l})
\end{equation}
Similarly, the inference probability, would be decomposed as
\begin{equation}
    q_\phi(z|x) = \prod_{l=1}^L q_\phi^{(l)}(z_l|x,z_{<l})
\end{equation}
where $q_\phi^{(l)}(z_l|x,z_{<l})$ is the encoder density of the $l$-th group.
Suitable (iterative) neural networks modules are used to sequentially compute the relevant statistics of these distributions, in terms of previous outputs. 

As an example of these architectures, the structure of NVAE will be detailed in Section~\ref{sec:nvae}.

The advantage of this approach is that it usually allows to work
with a larger number of latent variables, responsible for small and progressive adjustments of generated samples.

\subsection{Disentanglement}\label{sec:disentanglement}
Besides the task of generating new images, \cite{beta-vae17} and \cite{understanding-beta-vae18} noticed that VAEs can also be used to learn an efficient way to represent the data, with important applications in transfer learning and classification. 

To understand this phenomenon, suppose that there exists a set of \textit{true} generative factors $v = (v_1, \dots, v_S) \in \R{S}$ such that $p_{gt}(v|x) = \prod_{i=1}^S p_{gt}(v_i | x)$ (i.e. $v$ are conditionally independent given $x$) and that each $v_i$ encodes a meaningful feature of the data point $x$ generated by it. Under the assumption that $k \geq S$, the latent variables $z = (z_1, \dots, z_k)$ learnt during the training are a redundant representation of $v$ in a basis where the features are not disentangled. 
To learn an optimal latent representation of the input image $x$, it is necessary to train the network in such a way that $S$ coordinates of $z$ are related  to $v$, while the other $k - S$ coordinates can be used to improve the reconstruction of $x$, recovering the high frequency components that are missing in $v$. 

In $\beta$-VAE \cite{understanding-beta-vae18,beta-vae17}, this constraint is imposed by noting that in the ELBO function the prior distribution $p(z) = G(0, I)$ forces the decoder $q_\phi (z|x)$ to learn a vector $z$ where each variable is independent from each other. To improve disentanglement, 
we should hence induce the $D_{KL}$ term to be as small as possible, that can be achieved by augmenting the decoder variance $\gamma$ to be greater than 1. Unfortunately, since 
\begin{align*}
    \E{p_{gt}(x)} [ D_{KL} (q_\phi(z|x) || p(z)) ] = D_{KL} (q_\phi(z) || p(z)) + I_{q_\phi} (X; Z)
\end{align*}
where $I_{q_\phi}(X;Z)$ is the mutual information between $X$ and $Z$ with respect to the joint distribution $q_\phi(x, z) = q_\phi(z|x)p_{gt}(x)$, by pushing $D_{KL}(q_\phi(z|x)||p(z))$ to zero, the mutual information between $X$ and $Z$ is also minimized, reducing  the reconstruction efficiency of the network. 
This problem is addressed in \cite{disentangling_disentanglement,structured_disentangled_representation} where the ELBO is modified by adding more parameters with the intent to improve disentanglement without losing too much the performance. 

\section{Two-Stage VAE}\label{sec:two-stage}
To address the mismatch of aggregate posterior versus the expected prior, Bin Dai and David Wipf in \cite{TwoStage}, introduced the Two-Stage VAEs.

The idea behind this model is to train two different VAEs sequentially. The first VAE is used to learn a good representation $q_\phi(z|x)$ of the data in the latent space without guaranteeing  exactly  $q(z) = p(z)$, whereas  the second VAE should learn to sample from the true $q(z)$ without using the prior distribution $p(z)$.  A scheme of the implementation  follows 
(a detailed architectural 
description is given in Section~\ref{sec:arch}):

\begin{itemize}
	\item Given a data set $\D = \{ \x{1}, \dots, \x{N} \}$, train a VAE with a fixed latent dimension $k$, possibly small.  
	\item Generate latent samples $\mathcal{Z} = \{ z^{(1)}, \dots, z^{(N)} \}$ via $z^{(i)} \sim q_\phi(z|\x{i}), \ i=1, \dots N$. By design, these samples are distributed as $q_\phi(z) = \E{p_{gt}(x)} [q_\phi(z|x)]$, but likely not as $p(z) = G(0, I)$.
	\item Train a second VAE with parameters $(\theta', \phi')$ and latent variable $u \sim p(u) = G(0, I)$ of dimension $k$ to learn the distribution $q_\phi(z)$ with $\mathcal{Z}$ as the dataset.
	\item Sample new images by ancestral sampling, i.e. by first sampling $u \sim p(u)$, then generate a $z$ value by $p_{\theta '} (z|u)$ and finally $x \sim p_\theta(x|z)$.
\end{itemize}

The theoretical foundation of the Two-Stage VAE algorithm is well presented in \cite{TwoStage}. We  summarize here the main results. 
The two VAEs aim at separating the components of the ELBO loss function \eqref{eq:ELBO_split}, by suitably using the decoder variance $\gamma$.
Remarking that $p_{gt}(x)$ is the unknown data distribution which we desire to learn and that $p_\theta(x)=\E{q_\phi(z)} [p_\theta(x|z)]$ is the learnt distribution, we hope that $p_\theta (x) \approx p_{gt}(x) \> \forall x$. 

Unfortunately, this is not always possible. In fact, there is a critical distinction between the cases where the dimension of the data $d$ and the latent space dimension $k$ are equal, and the case where $d > k$. 

As a matter of facts, in the first case, it is possible to prove that, under suitable assumptions, for the optimal choice of the parameters $(\theta^*, \phi^*)$ it holds that $p_{\theta^*} (x) = p_{gt}(x)$ almost everywhere (i.e. VAEs strongly converges to the true distribution $p_{gt}(x)$).  In the second case,  only weak convergence, in the sense that $\int_A p_{\theta^*} (x) dx = \int_A p_{gt}(x) dx$ where $A$ is an open subset of $\R{d}$, can be proved (see Theorems 1 and 2 in \cite{TwoStage}).

In the first stage, since the ambient dimension is obviously greater than  the latent space dimension (i.e. $d>k$), for the previous results only a weak convergence is guaranteed; the parameter $\gamma$ is chosen in this case in order to get a good reconstruction (Theorem 4 in  \cite{TwoStage}). In the second stage by construction the data variable $z$ and its correspondent latent variable $u$ have the same dimension, hence the unknown distribution $q_\phi(z)$ is exactly identified by the VAE. As a consequence it is possible to sample directly from $q_\phi(z)$, without using the prior $p(z)$, thus bypassing the problem of mismatch between  the aggregate posterior and the prior distributions.

\section{Regularized VAE (RAE)}\label{sec:rae}
One of the most interesting variations of vanilla VAE is the work of Partha Ghosh and Mehdi S. M. Sajjadi \cite{deterministic}, where the authors tried to solve all the problems related to the classical VAE  by completely changing the the way  of approaching the problem. They pointed out that, in their typical implementation, VAEs can be seen as a regularized Autoencoder with Additive Gaussian Noise on the decoder input. 
In their work, the authors argued that noise injection in decoders input can be seen as a form of regularization, since it implicitly helps to smooth the function learnt by the network. 

To get a new insight to this problem, they took in consideration the distinct components of ELBO already introduced in (\ref{eq:ELBO_split}):

\begin{equation}
    \Loss{\theta, \phi} (x) := \underbrace{\E{q_{\phi}(z|x)} [ \log p_\theta(x|z) ]}_{:= \Loss{REC}(\theta, \phi)} - \gamma \underbrace{D_{KL}(q_\phi(z|x) || p(z))}_{:= \Loss{KL}(\phi)},
\end{equation}
where $\Loss{REC}$ is a term that measures the distance between the input and the reconstruction, whereas $\Loss{KL}$ is a regularization term that enforces the aggregate posterior to follow the prior distribution.

To show how $\Loss{KL} ( \phi)$ regularizes the loss, in \cite{deterministic}  the Constant-Variance VAEs (CV-VAEs) \cite{deterministic} have been investigated, where the encoder variance $\sigma^2_\phi(x)$ is fixed for every  $x \in \D$ and thus treated as an hyperparameter $\sigma^2$. In this situation, 
\begin{align}
	&\Loss{REC}(\theta, \phi) = - \E{q_\phi(z|x)} \Bigl[\frac{1}{2} || x - \mu_\theta(z) ||_2^2 \Bigr] \\
	&\Loss{KL}(\phi) = D_{KL}(q_\phi(z|x) || p(z)) = || \mu_\phi(x) ||_2^2 + C \\
	\label{eq:RAEloss}&\Loss{\theta, \phi}(x) = - \E{p_{gt}} \Bigl[ \E{q_\phi(z|x)} \Bigl[\frac{1}{2} || x - \mu_\theta(z) ||_2^2 \Bigr] - \gamma || \mu_\phi(x) ||_2^2 \Bigr].
\end{align}

We observe that the expression in \eqref{eq:RAEloss}  is a Mean Squared Error (MSE) with $L_2$ regularization on $\mu_\phi(x)$. \\

The authors proposed to substitute noise injection in the decoder input with an explicit regularization scheme in a classical CV-VAE. This is done by modifying the cost function $\Loss{\theta, \phi} = \E{p_{gt}(x)} [ \Loss{REC}(\theta, \phi) - \gamma \Loss{KL} (\phi) - \lambda \Loss{REG}(\theta) ]$ where $\Loss{REG}(\theta)$ is a regularizer for the decoder weights, while $\gamma, \lambda \geq 0$ are regularization parameters. \smallskip

Whereas $\Loss{REC} (\theta, \phi) = - \E{q_\phi(z|x)} [ \frac{1}{2} || x - \mu_\theta(z) ||_2^2 ]$ and $\Loss{KL}(\phi) = \frac{1}{2}|| z ||_2^2$ are fixed a priori by the CV-VAE architecture, $\Loss{REG}(\theta)$ needs to be defined. The choice  for $\Loss{REG}(\theta)$ identifies the specific kind of network. Ghosh and Sajjadi proposed three possible choices for $\Loss{REG}(\theta)$:

\begin{itemize}
	\item $L_2$-{\em Regularization}, where $\Loss{REG}(\theta) = || \theta ||_2^2$ is simply the weight decay on the decoder parameters.
	\item {\em Gradient Penalty}, where $\Loss{REG}(\theta) = || \nabla \mu_\theta(\mu_\phi(x))||_2^2$ bounds the gradient norm of the decoder with respect to its input, enforcing smoothness.
	\item {\em Spectral Normalization}, where each weight matrix $\theta_l$ in the decoder is normalized by an estimate of its largest singular value: $\theta_l^{SN} = \frac{\theta_l}{s(\theta_l)}$ (the estimate $s(\theta_l)$ can be easily obtained with one iteration of the power method). 
\end{itemize}

Moreover, they argued that removing noise injection from the decoder input prevents from knowing  the distribution  of latent variables, thus losing the generative ability of the network. 
They solved this problem by proposing an {\em ex-post density estimation}, where the distribution of the latent variables is learned a posteriori, by fitting $\mathcal{Z} = \{ z^{(i)}; z^{(i)} = \mu_\phi(\x{i}) \}$ with a  GMM  model $q_\delta(z)$ with a fixed number of components   and then sampling $z$ from $q_\delta(z)$ to generate new samples from $p_\theta(x|z)$. 
The generative model defined in this way is called {\em Regularized Autoencoder (RAE)}.

\section{Hierarchical Variational Autoencoder \label{sec:HVAE}}

To improve the quality of the generation in Variational Autoencoders, Kingma et al. \cite{autoregressive}  strengthened the inference network $q_\phi(z|x)$ with the powerful Normalizing Flows \cite{normalizing_flow} introduced by Rezende and Mohamed in 2015. The idea of Normalizing Flows is to begin with a latent variable $z_0$ sampled by a simple distribution $q_\phi(z_0 | x)$, and to iteratively construct more complex variables by applying transformations $z_t = f_t(z_{t-1})$ for $t = 1, \dots, T$. By observing that the $D_{KL}$ expression is:
\begin{equation}
    \label{eq:KL}
    D_{KL}(q_\phi(z_T|z_{<T}, x) || p(z_T)) = \E{q_\phi(z_T|z_{<T}, x)} \Bigl[ \log q_\phi(z_T|z_{<T}, x) - \log p(z_T)  \Bigr]
\end{equation}
its implementation requires the computation of the logarithm of $q_\phi(z_T|, z_{<T}, x)$.  If the functions $f_t(\cdot)$ are simple enough, it is possible to efficiently use them to compute $\log q_\phi(z_T| z_{<T}, x)$ as:
\begin{align}\label{eq:NF}
    \log q_\phi(z_T| z_{<T}, x) = \log q_\phi(z_0 | x) - \sum_{t=1}^T \log \det \Bigl| \frac{\partial f_t}{\partial z_{t-1}} \Bigr|
\end{align}
where $\frac{\partial f_t}{\partial z_{t-1}}$ is the Jacobian matrix of $f_t(z_{t-1})$  computed  by repeatedly applying the well known  change of variable theorem to the multi-variate random variable $z_T$ defined as:
\begin{align}
    z_T = f_T(f_{T-1}(\dots(f_1(z_0)) \dots))
\end{align}

\begin{figure}[h!]
    \centering
    \begin{tikzpicture}
    \draw (0.5, 1) rectangle (3.4, -4);
	\node[inputs] (X) at (0, 0) {$x$};
	\node[trapezium, trapezium angle=60, minimum width=25mm, fill=blue!30, draw, thick] (Encoder) at (2, 0) {$q_\phi(z_1|x)$};
 	\node[stats] (sigma) at (1.25, -1) {$\sigma_1^2$};
 	\node[stats] (mean) at (2.75, -1) {$\mu_1$};
	
  	\node[ops] (times) at (1.25, -2) {*};
  	\node[ops] (add) at (2, -2.5) {+};
 	
  	\node[sample] (epsil) at (0, -2) {$\epsilon$};
 	
  	\node[inputs] (z) at (2, -3.5) {$z_1$};
  	
  	\draw (3.6, 1) rectangle (6.4, -4);
  	\node[trapezium, trapezium angle=60, minimum width=25mm, fill=blue!30, draw, thick] (Encoder_2) at (5, 0) {$q_\phi(z_2|x)$};
 	\node[stats] (sigma_2) at (4.25, -1) {$\sigma_2^2$};
 	\node[stats] (mean_2) at (5.75, -1) {$\mu_2$};
 	
 	\node[ops] (times_2) at (4.25, -2) {*};
  	\node[ops] (add_2) at (5, -2.5) {+};
 	
  	\node[inputs] (z_2) at (5, -3.5) {$z_2$};
  	
    \draw (7.6, 1) rectangle (10.4, -4);
	\node[trapezium, trapezium angle=60, minimum width=25mm, fill=blue!30, draw, thick] (Encoder_T) at (9, 0) {$q_\phi(z_T|x)$};
 	\node[stats] (sigma_T) at (8.25, -1) {$\sigma_T^2$};
 	\node[stats] (mean_T) at (9.75, -1) {$\mu_T$};
 	
 	\node[ops] (times_T) at (8.25, -2) {*};
  	\node[ops] (add_T) at (9, -2.5) {+};
 	
  	\node[inputs] (z_T) at (9, -3.5) {$z_T$};

 	\draw[->] (X) -- (Encoder);
 	\draw[->] (Encoder) -- (sigma);
 	\draw[->] (Encoder) -- (mean);
    \draw[->] (sigma) -- (times);
    \draw[->] (times) .. controls (1.25, -2.5) .. (add);
    \draw[->] (mean) .. controls (2.75, -2.5) .. (add);
    \draw[->] (add) -- (z);
    \draw[->] (epsil) -- (times);
    
 	\draw[->] (Encoder_2) -- (sigma_2);
 	\draw[->] (Encoder_2) -- (mean_2);
 	\draw[->] (sigma_2) -- (times_2);
 	\draw[->] (mean_2) .. controls (5.75, -2.5) .. (add_2);
 	\draw[->] (z) .. controls (3.5, -3.5) and (3.5, -1) .. (times_2);
 	\draw[->] (times_2) .. controls (4.25, -2.5) .. (add_2);
 	\draw[->] (add_2) -- (z_2);
 	
 	\draw[->] (Encoder_T) -- (sigma_T);
 	\draw[->] (Encoder_T) -- (mean_T);
 	\draw[->] (sigma_T) -- (times_T);
 	\draw[->] (mean_T) .. controls (9.75, -2.5) .. (add_T);
 	\draw[->] (z_2) .. controls (7.5, -3.5) and (7.5, -1) .. (times_T);
 	\draw[->] (times_T) .. controls (8.25, -2.5) .. (add_T);
 	\draw[->] (add_T) -- (z_T);
 	
 	\node at (7, 0) {...};
 	\node at (7, -1) {...};
 	\node at (7, -3.5) {...};
 	
 	\node at (0.8, 0.8) {IAF};
 	\node at (3.9, 0.8) {IAF};
 	\node at (7.9, 0.8) {IAF};
    
\end{tikzpicture}
    \caption{A scheme of Inverse Autoregressive Flow. Each white box represents one iteration of equation \eqref{eq:iaf}, where $\mu_t, \sigma_t^2$ are generated by the encoder $q_\phi(z_t | x)$.}
    \label{fig:iaf}
\end{figure}
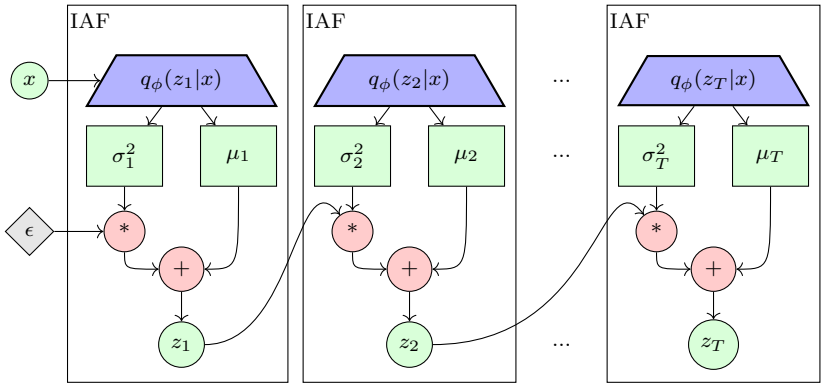

An interesting aspect concerning Normalizing Flows is that, under suitable assumptions, they are provably universal, in the sense defined in \cite{universality_NF}.
As already mentioned, the first successfully integration of Normalizing Flows in VAEs was by Kingma et al. in \cite{autoregressive}, where they introduced Inverse Autoregressive Flows (IAF). The idea was to define $f_t(z_{t-1})$ as a  simple affine function of the form:
\begin{align}
    z_t = f_t(z_{t-1}) = \mu_t + \sigma_t \odot z_{t-1} \quad \forall t = 1, \dots, T
    \label{eq:iaf}
\end{align}
where $z_0 \sim q_\phi(z_0 | x) = G(\mu_\phi(x), \sigma_\phi^2(x))$.\\
Figure \ref{fig:iaf} schematically represents the unrolling of equation \eqref{eq:iaf}.

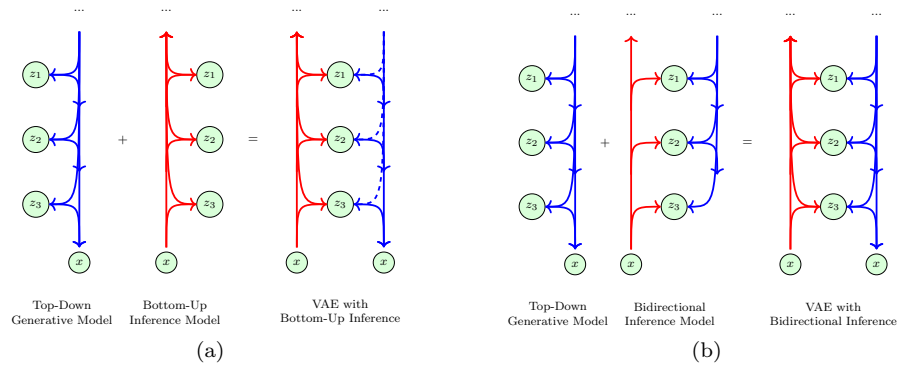
\begin{figure}[!ht]
\centering 
\subfloat[]{\resizebox{0.45 \textwidth}{!}{%
\begin{tikzpicture}
	\node[latent] (z1) at (0, 0) {$z_1$};
	\node[latent] (z2) at (0, -1.5) {$z_2$};
	\node[latent] (z3) at (0, -3) {$z_3$};
	
	\node[latent] (x) at (1, -4.35) {$x$};
	
	\draw[topdown] (1, 1) -- (1, -4);
	\draw[topdown] (1, 1) .. controls (1, 0) .. (z1);
	\draw[topdown] (1, 1) .. controls (1, -1.5) .. (z2);
	\draw[topdown] (1, 1) .. controls (1, -3) .. (z3);
	
	\draw[topdown] (z1) .. controls (1, 0) .. (1, -0.75);
	\draw[topdown] (z2) .. controls (1, -1.5) .. (1, -2.25);
	\draw[topdown] (z3) .. controls (1, -3) .. (1, -4);
	
	\node at (2, -1.5) {+};
	
	\node[latent] (z1) at (4, 0) {$z_1$};
	\node[latent] (z2) at (4, -1.5) {$z_2$};
	\node[latent] (z3) at (4, -3) {$z_3$};
	
	\node[latent] (x) at (3, -4.35) {$x$};
	
	\draw[bottomup] (3, -4) -- (3, 1);
	\draw[bottomup] (z1) .. controls (3, 0) .. (3, 1);
	\draw[bottomup] (z2) .. controls (3, -1.5) .. (3, 1);
	\draw[bottomup] (z3) .. controls (3, -3) .. (3,1);
	
	\draw[bottomup] (3, -0.75) .. controls (3, 0) .. (z1);
	\draw[bottomup] (3, -2.25) .. controls (3, -1.5) .. (z2);
	\draw[bottomup] (3, -4) .. controls (3, -3) .. (z3);
	
	\node at (5, -1.5) {=};
	
	\node[latent] (z1) at (7, 0) {$z_1$};
	\node[latent] (z2) at (7, -1.5) {$z_2$};
	\node[latent] (z3) at (7, -3) {$z_3$};
	
	\node[latent] (x) at (6, -4.35) {$x$};
	
	\draw[bottomup] (6, -4) -- (6, 1);
	\draw[bottomup] (z1) .. controls (6, 0) .. (6, 1);
	\draw[bottomup] (z2) .. controls (6, -1.5) .. (6, 1);
	\draw[bottomup] (z3) .. controls (6, -3) .. (6,1);
	
	\draw[bottomup] (6, -0.75) .. controls (6, 0) .. (z1);
	\draw[bottomup] (6, -2.25) .. controls (6, -1.5) .. (z2);
	\draw[bottomup] (6, -4) .. controls (6, -3) .. (z3);
	
	\node[latent] (x) at (8, -4.35) {$x$};
	
	\draw[topdown] (8, 1) -- (8, -4);
	\draw[topdown, dashed] (8, 1) .. controls (8, 0) .. (z1);
	\draw[topdown, dashed] (8, 1) .. controls (8, -1.5) .. (z2);
	\draw[topdown, ->, dashed] (8, 1) .. controls (8, -3) .. (z3);
	
	\draw[topdown] (z1) .. controls (8, 0) .. (8, -0.75);
	\draw[topdown] (z2) .. controls (8, -1.5) .. (8, -2.25);
	\draw[topdown] (z3) .. controls (8, -3) .. (8, -4);

	\node at (1, 1.5) {...};
	\node at (3, 1.5) {...};
	\node at (6, 1.5) {...};
	\node at (8, 1.5) {...};
	
	\node[align=center] at (0.6, -5.5) {Top-Down \\ Generative Model};
	\node[align=center] at (3.2, -5.5) {Bottom-Up \\ Inference Model};
	\node[align=center] at (7, -5.5) {VAE with \\ Bottom-Up Inference};
	
\end{tikzpicture} %
}} \hfill
\subfloat[]{\resizebox{0.45 \textwidth}{!}{%
\begin{tikzpicture}
	\node[latent] (z1) at (0, 0) {$z_1$};
	\node[latent] (z2) at (0, -1.5) {$z_2$};
	\node[latent] (z3) at (0, -3) {$z_3$};
	
	\node[latent] (x) at (1, -4.35) {$x$};
	
	\draw[topdown] (1, 1) -- (1, -4);
	\draw[topdown] (1, 1) .. controls (1, 0) .. (z1);
	\draw[topdown] (1, 1) .. controls (1, -1.5) .. (z2);
	\draw[topdown] (1, 1) .. controls (1, -3) .. (z3);
	
	\draw[topdown] (z1) .. controls (1, 0) .. (1, -0.75);
	\draw[topdown] (z2) .. controls (1, -1.5) .. (1, -2.25);
	\draw[topdown] (z3) .. controls (1, -3) .. (1, -4);
	
	\node at (1.7, -1.5) {+};
	
	\node[latent] (z1) at (3.3, 0) {$z_1$};
	\node[latent] (z2) at (3.3, -1.5) {$z_2$};
	\node[latent] (z3) at (3.3, -3) {$z_3$};
	
	\node[latent] (x) at (2.3, -4.35) {$x$};
	
	\draw[bottomup] (2.3, -4) -- (2.3, 1);
	
	\draw[bottomup] (2.3, -0.75) .. controls (2.3, 0) .. (z1);
	\draw[bottomup] (2.3, -2.25) .. controls (2.3, -1.5) .. (z2);
	\draw[bottomup] (2.3, -4) .. controls (2.3, -3) .. (z3);
	
	\draw[topdown] (4.3, 1) .. controls (4.3, -3) .. (z3);
	\draw[topdown] (4.3, 1) .. controls (4.3, 0) .. (z1);
	\draw[topdown] (4.3, 1) .. controls (4.3, -1.5) .. (z2);
	
	\draw[topdown] (z1) .. controls (4.3, 0) .. (4.3, -0.75);
	\draw[topdown] (z2) .. controls (4.3, -1.5) .. (4.3, -2.25);
	
	\node at (5, -1.5) {=};
	
	\node[latent] (z1) at (7, 0) {$z_1$};
	\node[latent] (z2) at (7, -1.5) {$z_2$};
	\node[latent] (z3) at (7, -3) {$z_3$};
	
	\node[latent] (x) at (6, -4.35) {$x$};
	
	\draw[bottomup] (6, -4) -- (6, 1);
	\draw[bottomup] (z1) .. controls (6, 0) .. (6, 1);
	\draw[bottomup] (z2) .. controls (6, -1.5) .. (6, 1);
	\draw[bottomup] (z3) .. controls (6, -3) .. (6,1);
	
	\draw[bottomup] (6, -0.75) .. controls (6, 0) .. (z1);
	\draw[bottomup] (6, -2.25) .. controls (6, -1.5) .. (z2);
	\draw[bottomup] (6, -4) .. controls (6, -3) .. (z3);
	
	\node[latent] (x) at (8, -4.35) {$x$};
	
	\draw[topdown] (8, 1) -- (8, -4);
	\draw[topdown] (8, 1) .. controls (8, 0) .. (z1);
	\draw[topdown] (8, 1) .. controls (8, -1.5) .. (z2);
	\draw[topdown, ->] (8, 1) .. controls (8, -3) .. (z3);
	
	\draw[topdown] (z1) .. controls (8, 0) .. (8, -0.75);
	\draw[topdown] (z2) .. controls (8, -1.5) .. (8, -2.25);
	\draw[topdown] (z3) .. controls (8, -3) .. (8, -4);

	\node at (1, 1.5) {...};
	\node at (2.3, 1.5) {...};
	\node at (4.3, 1.5) {...};
	\node at (6, 1.5) {...};
	\node at (8, 1.5) {...};
	
	\node[align=center] at (0.6, -5.5) {Top-Down \\ Generative Model};
	\node[align=center] at (3.2, -5.5) {Bidirectional \\ Inference Model};
	\node[align=center] at (7, -5.5) {VAE with \\ Bidirectional Inference};
	
\end{tikzpicture} %
}}
\caption{Diagrams that schematically represents  Hierarchical VAE in two different configurations:  Bottom-Up Inference (a) and  Bidirectional Inference (b).}
\label{fig:hierarchical_architecture}
\end{figure}

We highlight that the IAF   introduces a natural order in the latent variables. For this reason, we will refer to this kind of models as Hierarchical Variational Autoencoder (HVAE). In this paradigm, we will refer to each $z_t$ as a group of latent variables, and we will collect the set of all groups in a vector $z = (z_0, \dots, z_T)$ where the variables are written in the order defined above. \\
If we distinguish between the encoder (inference) network $q_\phi(z|x)$ and the decoder (generative) network, we need to choose if the ordering of latent variables is the same in the two parts of the network (\textit{bottom-up inference}) , or if it is reversed (\textit{bidirectional inference}) as shown in Figure \ref{fig:hierarchical_architecture}.

As it is clear from Figure \ref{fig:hierarchical_architecture}, in bottom-up inference the image $x \in \R{d}$ is encoded to $z = (z_1, \dots, z_T)$ independently from the prior $p(z) = \prod_{t=1}^T p(z_t|z_{<t})$; in the generative phase the image is reconstructed by taking $z_T$ as  the final output of the encoder, and then sampling each $z_t$, $t = T-1, \dots, 0$ from the prior distribution independently from $q_\phi(z_t|x)$ (i.e. the encoder and decoder are independent from each other). We underline that  this fact makes the bottom-up inference training  unstable. 

Conversely, in bidirectional inference, the process of generating latent variables is shared between the two parts of the network, which  makes the training easier, 
although the design of the network is a bit more difficult.\smallskip

Since the results of vanilla IAF are not competitive with the state-of-art, we will not use them in our future analysis (see the original paper for more information), whereas  we will focus our experimental results on  two powerful variants of IAF, making use of bidirectional inference and residual blocks to generate high quality images.

\section{Experimental setting}\label{sec:setting}
For each variant of Variational Autoencoder discussed in the previous sections, we provide an 
original implementation in TensorFlow 2, and a set of detailed benchmarks on traditional
datasets, such as MNIST, Cifar10 and CelebA. The specific architectures which have been tested 
are described in the following. All models have been compared using 
standard metrics, assessing both their energy consumption through the number of floating point operations (see Section~\ref{sec:flops}), and their performance
via the so called Frech\`et Inception Distance \cite{FID}, briefly discussed in Section~\ref{sec:FID}. 
Numerical results are given in Section~\ref{sec:tables}, along with examples of reconstructed and generated images.

\subsection{Green AI and FLOPS}\label{sec:flops}
The paradigm of Green AI \cite{GreenAI} is meant to raise the attention on the computational efficiency of neural models, encouraging a reduction in
the amount of resources required for their training and deployment. 
This concept is not so trivial as it seems; in fact, most of traditional
AI research (referred to as Red AI, in this con) targets accuracy rather than efficiency, exploiting massive computational power, and resulting in rapidly escalating costs; this trend is not sustainable for various reasons, it is environmentally unfriendly \cite{lacoste2019quantifying}, socially not inclusive and inefficient.

The computation of 
floating point operations (FLOPS) was advocated in \cite{GreenAI} as a measure of the efficiency of models;
the main advantages of this measure are that it is hardware independent
and has a direct (even if not precise) correlation with the running time of the model \cite{canziani2017analysis}. There are also known problems related to FLOPs, mostly related to the fact that memory access
time can be a more dominant factor in real implementations (see the ``Trap of FLOPs" discussion in \cite{trap_of_flops}).

So, while we shall adopt FLOPS for our comparison, we shall also 
investigate performance through more traditional tools, like Tensorboard, 
also in order to gain confidence on the reliability of FLOPs-based assesments.

\subsection{Frech\`et Inception Distance}\label{sec:FID}
To test the quality of the generator, we should compare the probability distribution of
generated vs  real images. Unfortunately, the dimension of the feature space is 
typically too large to allow a direct, significant comparison; moreover, in the case of images, adjacent pixels are highly correlated, reducing their statistical relevance. 
The main idea behind 
the so called Frech\`et Inception Distance (FID) \cite{FID} is to use, instead of raw data,
their internal representations generated by some third party, agnostic network. In the
case of FID, the Inception v3 network \cite{InceptionV3} trained on Imagenet is used to this purpose; 
Inception is usually preferred over other models due to the limited amount 
of preprocessing performed on input images (images are rescaled in the interval [-1,1], sample wise). The activations that are traditionally used
are those relative to the last pooling layer, with a dimension of 2048 features.

Given the activations $a_1$ and $a_2$, relative to real and generated images, and called $\mu_i, i=1,2$ and $C_i, i=1,2$ their empirical mean and covariance matrix, respectively, the  Fr\`echet Distance between $a_1$ and $a_2$ is defined as:
\begin{equation}\label{eq:FID}
FID(a_1,a_2) = ||\mu_1 – \mu_2||^2 + Tr(C_1 + C_2 – 2(C_1*C_2)^{\frac{1}{2}})
\end{equation}
where we indicate with $Tr$ the trace of a matrix.  

A problem of FID, is that it is extremely sensible to a number of different factors
elencated below.
\begin{enumerate}
\item the weights of the Inception network. The checkpoint that is traditionally used is
the \verb+inception_v3_2016_08_28/inception_v3.ckpt+ downloaded from TF-Slim's pre-trained models, also available through Tensorflow-HUB. These weights were originally obtained by 
training on the ILSVRC-2012-CLS dataset for image classification ("Imagenet"). 
\item The dimension of the datasets of real/generated images to be compared. Traditionally,
sets of 10K images are compared; typically, the FID score is inversely proportional to this
dimension.
\item The dimension of input images fed to Inception. Inception may work with images of
arbitrary size (larger than $75\times 75$), however the ``canonical'' input dimension is 
$299 \times 299$. Again, varying the dimension may result in very different scores.
\item The resizing algorithm. Images must be resized to bring them to the expected input
dimension of $299 \times 299$; as observed in \cite{balancing}, the FID score is quite sensible
to the algorithm used to this aim, and in particular to the employed modality: nearest neighbour, bilinear interpolation, cubic interpolation, \dots. The default, is usually bilinear interpolation, being a good compromise between efficiency and quality.
\end{enumerate}
Unfortunately, articles in the literature are not always fully transparent on the previous
points, that may explain some discrepancies and the difficulty one frequently faces in replicating results. 

All our experiments have been conducted with "defaults" values: the standard Inception checkpoint \verb+inception_v3_2016_08_28/inception_v3.ckpt+, 10000 images of dimension
$299\times 299$, rescaled by means of bilinear interpolation.

Let us finally observe that, in the case of VAE, it is customary to measure both 
the FID score for reconstructed images ($FID_{rec}$) and the FID score for generated
images ($FID_{gen}$). The former one is usually reputed to be a lower bound for the
latter, no matter what help we may provide to the generator during the sampling face.

\subsection{Architectures overview}\label{sec:arch}
In  this section, we provide detailed descriptions of the several different neural networks
architectures we have been dealing with, each one inspired by a different article. 
For each of them, different possible configurations
have been investigated, varying the number and dimension of layers, as well as
the learning objectives. Moreover, since some of the techniques considered
are not dependent from  the encoder/decoder structure, we also tested a mix of different architectures, hyperparameters configurations, and optimization objectives.

\subsubsection{Vanilla Convolutional VAE}
\begin{figure}
    \centering
    \resizebox{\linewidth}{!}{%
\begin{tikzpicture}
\tikzstyle{connection}=[ultra thick,every node/.style={sloped,allow upside down},draw=\edgecolor,opacity=0.7]
\tikzstyle{copyconnection}=[ultra thick,every node/.style={sloped,allow upside down},draw={rgb:blue,4;red,1;green,1;black,3},opacity=0.7]

\pic[shift={(0,0,0)}] at (0,0,0) 
    {Box={
        name=conv0,
        caption= Input,
        xlabel={{3, }},
        zlabel=32,
        fill=\ConvColor,
        height=32,
        width=1.5,
        depth=32
        }
    };

\pic[shift={ (0,0,0) }] at (conv0-east) 
{Box={
		name=pool0,
		caption= ,
		fill=\PoolColor,
		opacity=0.5,
		height=16,
		width=1,
		depth=16
	}
};

\pic[shift={(1,0,0)}] at (conv0-east) 
    {Box={
        name=conv1,
        caption= Conv1,
        xlabel={{128, }},
        zlabel=16,
        fill=\ConvColor,
        height=16,
        width=3,
        depth=16
        }
    };

\draw [connection]  (conv0-east)    -- node {\midarrow} (conv1-west);

\pic[shift={ (0,0,0) }] at (conv1-east) 
    {Box={
        name=pool1,
        caption= ,
        fill=\PoolColor,
        opacity=0.5,
        height=8,
        width=1,
        depth=8
        }
    };

\pic[shift={(1,0,0)}] at (pool1-east) 
    {Box={
        name=conv2,
        caption= Conv2,
        xlabel={{256, }},
        zlabel=8,
        fill=\ConvColor,
        height=8,
        width=6,
        depth=8
        }
    };

\draw [connection]  (pool1-east)    -- node {\midarrow} (conv2-west);

\pic[shift={ (0,0,0) }] at (conv2-east) 
    {Box={
        name=pool2,
        caption= ,
        fill=\PoolColor,
        opacity=0.5,
        height=4,
        width=1,
        depth=4
        }
    };

\pic[shift={(1,0,0)}] at (pool2-east) 
    {Box={
        name=conv3,
        caption= Conv3,
        xlabel={{512, }},
        zlabel=4,
        fill=\ConvColor,
        height=4,
        width=10,
        depth=4
        }
    };

\draw [connection]  (pool2-east)    -- node {\midarrow} (conv3-west);

\pic[shift={ (0,0,0) }] at (conv3-east) 
{Box={
		name=pool3,
		caption= ,
		fill=\PoolColor,
		opacity=0.5,
		height=2,
		width=1,
		depth=2
	}
};

\pic[shift={(1,0,0)}] at (pool3-east) 
    {Box={
        name=conv4,
        caption= Conv4,
        xlabel={{1024, }},
        zlabel=2,
        fill=\ConvColor,
        height=2,
        width=15,
        depth=2
        }
    };

\draw [connection]  (pool3-east)    -- node {\midarrow} (conv4-west);

\pic[shift={(1,0,2)}] at (conv4-east) 
    {Box={
        name=Fc1,
        caption=FC,
        xlabel={{, }},
        zlabel=128,
        fill=\FcColor,
        height=1,
        width=1,
        depth=10
        }
    };

\draw [connection]  (conv4-east)    -- node {\midarrow} (Fc1-west);

\pic[shift={(1,0,-2)}] at (conv4-east) 
{Box={
		name=Fc2,
		caption=,
		xlabel={{, }},
		zlabel=128,
		fill=\FcColor,
		height=1,
		width=1,
		depth=10
	}
};

\draw [connection]  (conv4-east)    -- node {\midarrow} (Fc2-west);

\pic[shift={(2,0,0)}] at (conv4-east) 
    {Box={
        name=Fc3,
        caption= $z$,
        xlabel={{,}},
        zlabel=128,
        fill=\FcColor,
        height=1,
        width=1,
        depth=10
        }
    };

\draw [connection]  (Fc1-east)    -- node {\midarrow} (Fc3-west);
\draw [connection]  (Fc2-east)    -- node {\midarrow} (Fc3-west);

\pic[shift={(1,0,0)}] at (Fc3-west) 
{Box={
		name=conv2t,
		caption= Conv2T,
		xlabel={{512, }},
		zlabel=8,
		fill=\ConvColor,
		height=8,
		width=10,
		depth=8
	}
};

\pic[shift={ (0,0,0) }] at (conv2t-east) 
{Box={
		name=unpool3,
		caption= ,
		fill=\PoolColor,
		opacity=0.5,
		height=8,
		width=1,
		depth=8
	}
};

\pic[shift={(1,0,0)}] at (unpool3-east) 
{Box={
		name=conv1t,
		caption= Conv1T,
		xlabel={{256, }},
		zlabel=16,
		fill=\ConvColor,
		height=16,
		width=6,
		depth=16
	}
};

\pic[shift={ (0,0,0) }] at (conv1t-east) 
{Box={
		name=unpool4,
		caption= ,
		fill=\PoolColor,
		opacity=0.5,
		height=16,
		width=1,
		depth=16
	}
};

\pic[shift={(1,0,0)}] at (unpool4-east) 
{Box={
		name=output,
		caption= Output,
		xlabel={{3, }},
		zlabel=32,
		fill=\ConvColor,
		height=32,
		width=1.5,
		depth=32
	}
};

\draw [connection]  (Fc3-east)    -- node {\midarrow} (conv2t-west);
\draw [connection]  (unpool3-east)    -- node {\midarrow} (conv1t-west);
\draw [connection]  (unpool4-east)    -- node {\midarrow} (output-west);

\end{tikzpicture}%
}
    \caption{Graphical representation of the Vanilla VAE architecture. The yellow, orange and green  boxes represent convolutional, downsampling and dense layers, respectively. }
    \label{fig:cnn_vanilla_vae}
\end{figure}
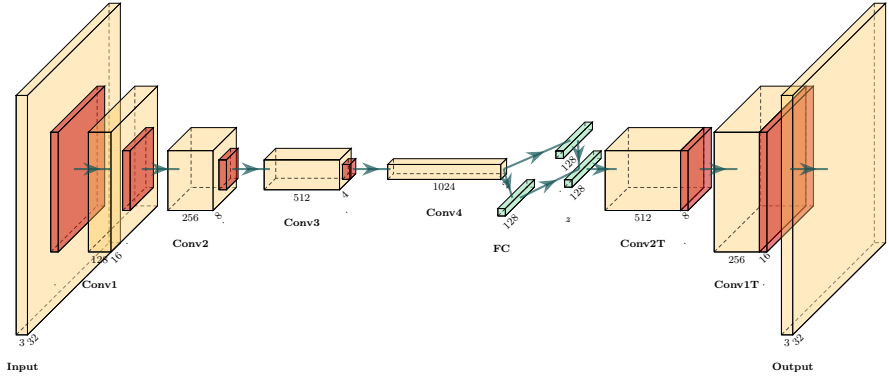

In our first experiment we followed the same structure of \cite{deterministic}, which is a simple CNN architecture where we doubled the number of channels for each Convolution, and we down-sampled the spatial dimension by 2 (see Figure \ref{fig:cnn_vanilla_vae}). \\
The encoder is structured as follows. In the first layer, the input image of dimension $(d, d, 3)$ (where $d = 32$ and   $d = 64$  in CIFAR10 and CelebA, respectively) was passed through a convolutional layer with 128 channels and stride equals 2, to obtain 128 images of dimension $(d/2, d/2)$. This operation is repeated for $256, 512, 1024$ channels. The result is flattened and passed through two Dense layers to obtain the mean and the variance of the latent variables. \\
The decoder has the same structure of the encoder, with Transposed convolutions and Upsample layers. \\
Each convolutional filter has kernel size  4 and ReLU activation function, except for the last layer of the decoder, where we used a sigmoid activation to ensure that the output is in the range $[0, 1]$. 

\subsubsection{Resnet-like}
The Resnet-like architecture was adopted in \cite{TwoStage}. 
The main difference of this newtork with respect to the Vanilla VAE is that, before
downsampling, the input is processed by a so called {\em Scale Block},
that is just a sequence of Residual Blocks. In turn, a Residual Block is an alternated sequence of BatchNormalization and Convolutional layers (with unit stride), intertwined with residual connections. 
The number of Scale Blocks at each scale of the image pyramid,
the number of Residual Blocks inside each Scale Block, and the
number of convolutions inside each Residual Block are 
user configurable hyperparameters.

\begin{figure}[ht]
\begin{center}
\begin{tabular}{cc}
\raisebox{1cm}{\includegraphics[width=.4\textwidth]{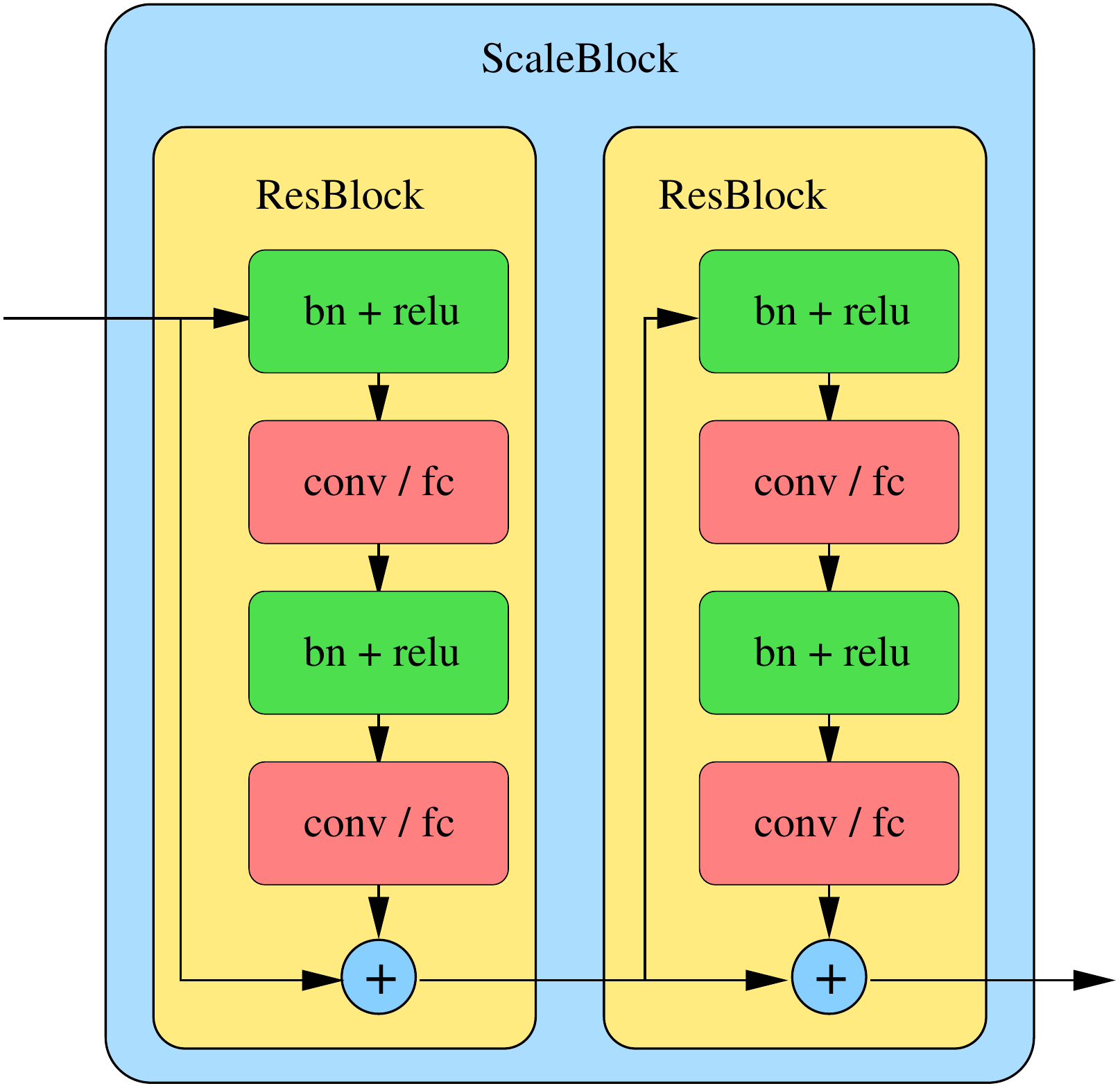}}\hspace*{.45cm}&\hspace{.4cm}
\includegraphics[width=.4\textwidth]{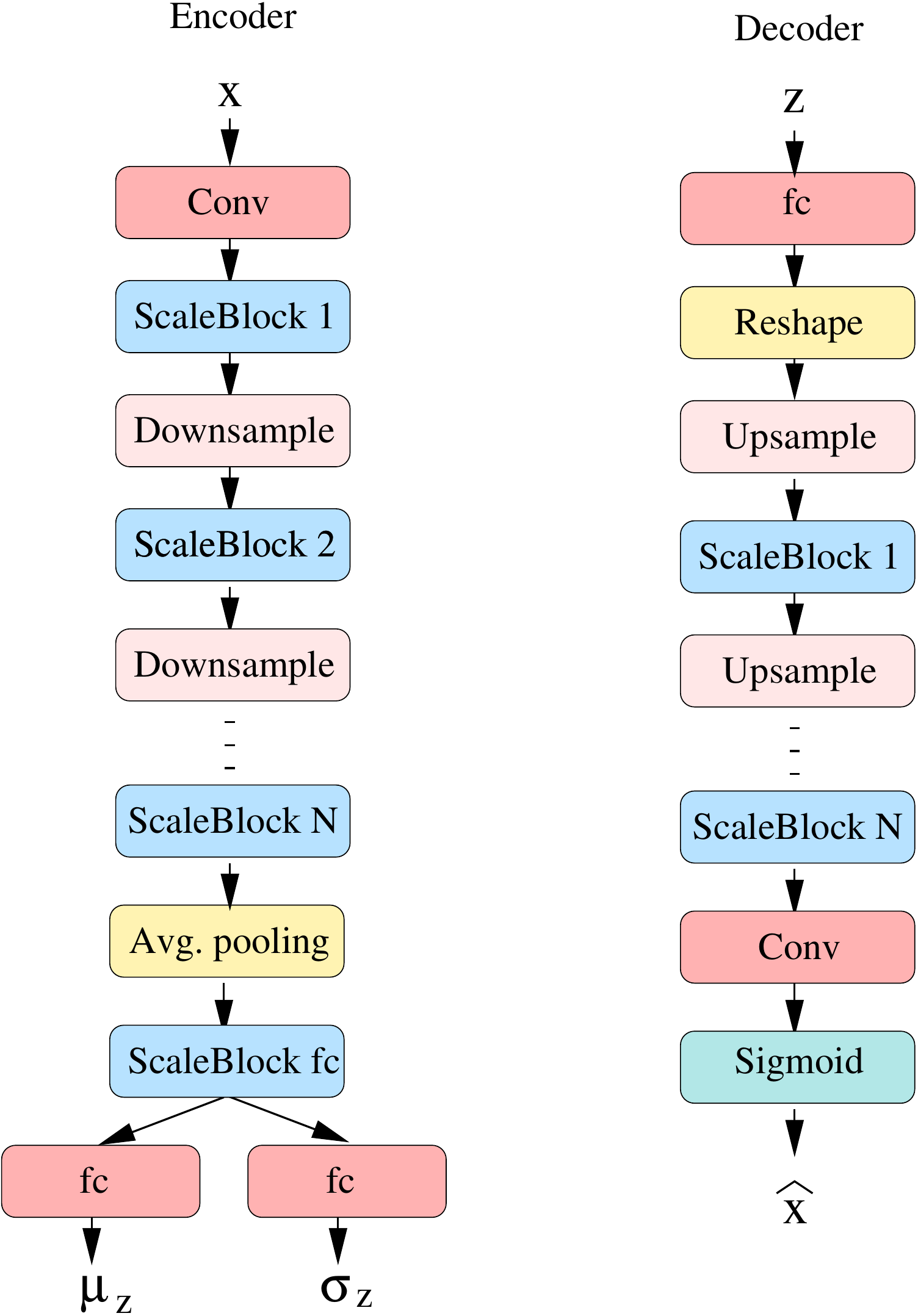} \\
(a) & (b)
\end{tabular}
\caption{(a) Scale Block. The Scale Block is used to process features at a given scale; it is a sequence of Residual Blocks intertwined with residual connections. A Residual Block is an alternation of batchnormalization layers, rectified linear units and convolutions. (b) The input is progressively downsampled via convolutions with stride 2, intermixed by Scale Blocks; at a given scale, a global average pooling layer extract features that are further processed via dense layers to compute mean and variance for latent variables. The decoder
is essentially symmetric.}
\label{fig:Resnet}
\end{center}
\end{figure}

In the encoder, at the end of the last Scale Block, a global
average level extracts spatial agnostic features. These are
first passed through a so called {Dense Block} (similar to
a Residual Block but with dense layers instead of convolutions),
and finally used to synthesize mean and variance for latent variables.

The decoder first maps the internal encoding $z$ to a small
map of dimension $4\times 4 \times base\_dim$ via a dense layer
suitably reshaped. This is then up-sampled to the final expected
dimension, inserting Scale Blocks at each scale.

\subsubsection{Two-Stage VAE}
To check in what extent the Two-Stage VAE improve the generation ability of a Variational Autoencoder, we tried to fit a second stage to every model we tested, following the architecture described in the following and graphically represented in Figure \ref{fig:Resnet}. \\
The encoder in the second stage model in composed of a couple of Dense layers of dimension 1536 and ReLU activation function, followed by a concatenation with the Input of the model and then by another Dense layer to obtain the latent representation $u$ with the same dimension of $z$, following what's described in Section \ref{sec:two-stage}. The decoder has exactly the same structure of the encoder. \\
As already described, we used the \textit{cosine similarity} as the reconstruction part of the ELBO objective function. \\
We observed that, to improve the quality of the generation, the second stage should be trained for a large number of epochs.

\subsubsection{Convolutional RAE}
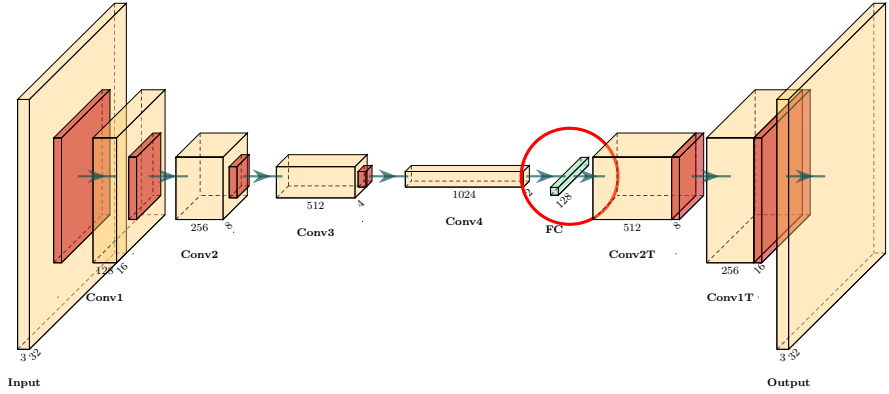
\begin{figure}
    \centering
    \resizebox{\linewidth}{!}{%
\begin{tikzpicture}
\tikzstyle{connection}=[ultra thick,every node/.style={sloped,allow upside down},draw=\edgecolor,opacity=0.7]
\tikzstyle{copyconnection}=[ultra thick,every node/.style={sloped,allow upside down},draw={rgb:blue,4;red,1;green,1;black,3},opacity=0.7]

\pic[shift={(0,0,0)}] at (0,0,0) 
    {Box={
        name=conv0,
        caption= Input,
        xlabel={{3, }},
        zlabel=32,
        fill=\ConvColor,
        height=32,
        width=1.5,
        depth=32
        }
    };

\pic[shift={ (0,0,0) }] at (conv0-east) 
{Box={
		name=pool0,
		caption= ,
		fill=\PoolColor,
		opacity=0.5,
		height=16,
		width=1,
		depth=16
	}
};

\pic[shift={(1,0,0)}] at (conv0-east) 
    {Box={
        name=conv1,
        caption= Conv1,
        xlabel={{128, }},
        zlabel=16,
        fill=\ConvColor,
        height=16,
        width=3,
        depth=16
        }
    };

\draw [connection]  (conv0-east)    -- node {\midarrow} (conv1-west);

\pic[shift={ (0,0,0) }] at (conv1-east) 
    {Box={
        name=pool1,
        caption= ,
        fill=\PoolColor,
        opacity=0.5,
        height=8,
        width=1,
        depth=8
        }
    };

\pic[shift={(1,0,0)}] at (pool1-east) 
    {Box={
        name=conv2,
        caption= Conv2,
        xlabel={{256, }},
        zlabel=8,
        fill=\ConvColor,
        height=8,
        width=6,
        depth=8
        }
    };

\draw [connection]  (pool1-east)    -- node {\midarrow} (conv2-west);

\pic[shift={ (0,0,0) }] at (conv2-east) 
    {Box={
        name=pool2,
        caption= ,
        fill=\PoolColor,
        opacity=0.5,
        height=4,
        width=1,
        depth=4
        }
    };

\pic[shift={(1,0,0)}] at (pool2-east) 
    {Box={
        name=conv3,
        caption= Conv3,
        xlabel={{512, }},
        zlabel=4,
        fill=\ConvColor,
        height=4,
        width=10,
        depth=4
        }
    };

\draw [connection]  (pool2-east)    -- node {\midarrow} (conv3-west);

\pic[shift={ (0,0,0) }] at (conv3-east) 
{Box={
		name=pool3,
		caption= ,
		fill=\PoolColor,
		opacity=0.5,
		height=2,
		width=1,
		depth=2
	}
};

\pic[shift={(1,0,0)}] at (pool3-east) 
    {Box={
        name=conv4,
        caption= Conv4,
        xlabel={{1024, }},
        zlabel=2,
        fill=\ConvColor,
        height=2,
        width=15,
        depth=2
        }
    };

\draw [connection]  (pool3-east)    -- node {\midarrow} (conv4-west);

\pic[shift={(1,0,0)}] at (conv4-east) 
    {Box={
        name=Fc3,
        caption= FC,
        xlabel={{,}},
        zlabel=128,
        fill=\FcColor,
        height=1,
        width=1,
        depth=10
        }
    };

\draw[color=red, line width=0.8mm] (12.8,0) circle (35pt);
\draw [connection]  (conv4-east)    -- node {\midarrow} (Fc3-west);

\pic[shift={(1,0,0)}] at (Fc3-west) 
{Box={
		name=conv2t,
		caption= Conv2T,
		xlabel={{512, }},
		zlabel=8,
		fill=\ConvColor,
		height=8,
		width=10,
		depth=8
	}
};

\pic[shift={ (0,0,0) }] at (conv2t-east) 
{Box={
		name=unpool3,
		caption= ,
		fill=\PoolColor,
		opacity=0.5,
		height=8,
		width=1,
		depth=8
	}
};

\pic[shift={(1,0,0)}] at (unpool3-east) 
{Box={
		name=conv1t,
		caption= Conv1T,
		xlabel={{256, }},
		zlabel=16,
		fill=\ConvColor,
		height=16,
		width=6,
		depth=16
	}
};

\pic[shift={ (0,0,0) }] at (conv1t-east) 
{Box={
		name=unpool4,
		caption= ,
		fill=\PoolColor,
		opacity=0.5,
		height=16,
		width=1,
		depth=16
	}
};

\pic[shift={(1,0,0)}] at (unpool4-east) 
{Box={
		name=output,
		caption= Output,
		xlabel={{3, }},
		zlabel=32,
		fill=\ConvColor,
		height=32,
		width=1.5,
		depth=32
	}
};

\draw [connection]  (Fc3-east)    -- node {\midarrow} (conv2t-west);
\draw [connection]  (unpool3-east)    -- node {\midarrow} (conv1t-west);
\draw [connection]  (unpool4-east)    -- node {\midarrow} (output-west);

\end{tikzpicture}%
}
    \caption{Graphical representation of the RAE architecture. The yellow, orange and green  boxes represent convolutional, downsampling and dense layers, respectively. The red circle underlines the sole architectural difference between our implementation of VanillaVAE and RAE, i.e. the fact that in the latter, the latent space is composed by a single Dense layer that directly encodes to $z$, while in VanillaVAE the encoding is performed by a couple of Dense layers that represents the mean and the variance of a Gaussian distribution. }
    \label{fig:cnn_rae}
\end{figure}

In our implementation of RAE, we followed exactly the same structure as in Convolutional Vanilla VAE, with the sole difference that, in RAEs, the latents space is composed of just one fully connected layer representing the variable $z$ (see Figure \ref{fig:cnn_rae}). \\
In our tests, we only compared $L_2$ and GP regularization, with parameter $\lambda$ heuristically computed to achieve the best performance.

\subsubsection{NVAE}\label{sec:nvae}
\begin{figure}[ht]
\begin{center}
\resizebox{0.6 \textwidth}{!}{%
\begin{tikzpicture}
\node[latent] (z1) at (1.5, 1.25) {$z_1$};
\node[latent] (z2) at (1.5, -0.75) {$z_2$};
\node[latent] (z3) at (1.5, -2.75) {$z_3$};

\node[latent] (x) at (0, -5) {$x$};
\node[latent] (x_rec) at (3, -5) {$\hat{x}$};

\node[sample] (er1) at (0, -3.5) {r};
\node[sample] (dr1) at (3, -3.5) {r};
\node[ops] (add1) at (3, -2.75) {+};

\node[sample] (er2) at (0, -1.5) {r};
\node[sample] (dr2) at (3, -1.5) {r};
\node[ops] (add2) at (3, -0.75) {+};

\node[sample] (er3) at (0, 0.5) {r};
\node[sample] (dr3) at (3, 0.5) {r};

\draw[bottomup] (x) -- (er1);
\draw[bottomup] (er1) -- (er2);
\draw[bottomup] (er2) -- (er3);

\draw[topdown] (dr3) -- (add2);
\draw[topdown] (dr2) -- (add1);
\draw[topdown] (dr1) -- (x_rec);

\draw[topdown] (z3) -- (add1);
\draw[topdown] (z2) -- (add2);

\draw[bottomup] (er3) .. controls (0, 1.25) .. (z1);

\node[ops] (add3) at (1.5, -1.8) {+};
\node[ops] (add4) at (1.5, 0.2) {+};

\draw[bottomup] (0.1, 0.2) -- (add4);
\draw[bottomup] (0.1, -1.8) -- (add3);

\draw[topdown] (2.9, 0.2) -- (add4);
\draw[topdown] (2.9, -1.8) -- (add3);

\draw[topdown] (add4) -- (z2);
\draw[topdown] (add3) -- (z3);

\node[ops] (add5) at (3, 1.25) {+};
\draw[topdown] (z1) -- (add5);

\node[stats] (h) at (3, 2.5) {h};
\draw[topdown] (h) -- (add5);

\node[latent] (z1) at (6.5, 1.25) {$z_1$};
\node[latent] (z2) at (6.5, -0.75) {$z_2$};
\node[latent] (z3) at (6.5, -2.75) {$z_3$};

\node[latent] (x_rec) at (8, -5) {$\hat{x}$};

\node[sample] (dr1) at (8, -3.5) {r};
\node[ops] (add1) at (8, -2.75) {+};

\node[sample] (dr2) at (8, -1.5) {r};
\node[ops] (add2) at (8, -0.75) {+};

\node[sample] (dr3) at (8, 0.5) {r};

\draw[topdown] (dr3) -- (add2);
\draw[topdown] (dr2) -- (add1);
\draw[topdown] (dr1) -- (x_rec);

\draw[topdown] (z3) -- (add1);
\draw[topdown] (z2) -- (add2);

\node[ops] (add5) at (8, 1.25) {+};
\draw[topdown] (z1) -- (add5);

\node[stats] (h) at (8, 2.5) {h};
\draw[topdown] (h) -- (add5);

\draw[topdown] (dr3) .. controls (6.5, 0.5) .. (z2);
\draw[topdown] (dr2) .. controls (6.5, -1.5) .. (z3);

\node at (1.5, -6) {(a)};
\node at (7, -6) {(b)};
\end{tikzpicture} %
}
\caption{The whole NVAE architecture (a) and a focus on its decoder (b).  }
\label{fig:nvae}
\end{center}
\end{figure}
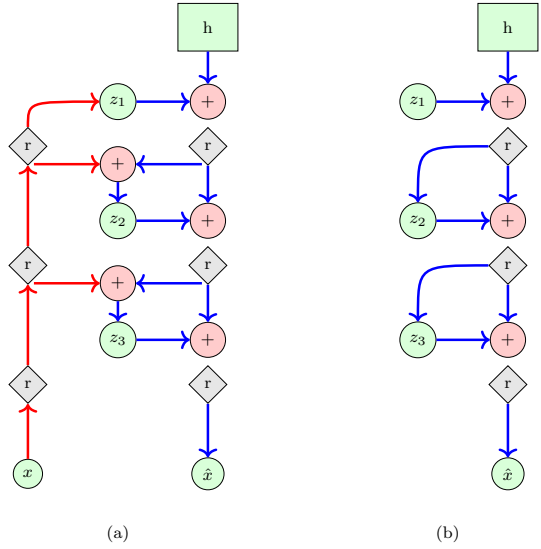

The model is organized in a bottom-up inference network and a
top-down generative network (see Figure~\ref{fig:nvae}). Each one of two networks is composed
by a hierarchy of modules at different {\em scales}. Each scale is composed by  {\em groups} of sequential (residual) blocks. 

During generation, each module computes from the current input $h_l$
a prior $p(z_l|h_l)$ ($h_l$ depends from  $z_{<l}$): after sampling
from this prior, the result is combined in some way with the current
input $h_l$, the two informations are processed together and passed 
to the next module.

During inference, we extract the latent representation at stage $l$ by 
synthesizing a mean and a standard deviation for $q(z_l|x,h_l)$: since 
this information depends from $h_l$, we expect to provide additional 
information, not already available by previous latent encodings. 
Moreover, the computation of $h_l$, is done by the top-down network, that is
hence a sub-network of the inference network. During training, 
both networks are trained together.

Each network has a hierarchical organization at
different {\em scales}. Each scale is composed by {\em groups} of Blocks. 

Both Encoder Blocks (EB) and Decoder Blocks (DB) have similar architectures, and are essentially composed by an alternated sequence of BatchNormalization and Convolutional layers, separated by non linear activation layers, and intermixed with residual connections (so, very similar to the Scale Block discussed in the previous section). A few technical novelties are however introduced by the authors:
\begin{itemize}
    \item the recent {\em Swish} activation function 
    $f(u) = \frac{u}{1+e^{-u}}$ \cite{swish} is used instead of Relu, Elu,
    or other more traditional choices;
    \item a Squeeze-and-Excitation (SE) layer \cite{SE}
    is added at the end of each block;
    \item a moderate use of depthwise separable 
    convolutions \cite{xception} is deployed in order to reduce the number of parameters of the network.
\end{itemize}

Table~\ref{tab:nvae-hyperparameters}  gives a summary of hyperparameters used in training NVAE on the datasets addressed in this article, borrowed from \cite{NVAE}. $D^2$
indicates a latent variable with the spatial dimensions of $D \times D$. As an example, the MNIST model
consists of two scales: in the first one, we
have five groups of $4 \times 4 \times 20$-dimensional latent variables: in the second one, we have 10 groups of 
$8 \times 8 \times 20$-dimensional variables.

\begin{table}[ht]
    \centering
    \begin{tabular}{lccc}
         hyperparameter &  MNIST & Cifar10 & CelebA \\\hline
         input size & $28\times28$ &  $32\times32$ &  $64\times64$ \\\hline
         epochs & 400 & 400 & 90 \\\hline
         batchsize & 200 & 32 & 16 \\\hline
         normalizing flows & 0 & 2 & 2\\\hline
         scales & 2 & 1 & 3 \\\hline
         groups per scale & 5,10 & 30 & 5,10,20 \\\hline
         spatial dims of z per scale & $4^2,8^2$ & $16^2$ & $8^2,16^2,32^2$ \\\hline
         channel dims of z & 20 & 20 & 20 \\\hline
         initial channels in enc. & 32 & 128 & 64 \\\hline
         residual cells per group & 1 & 2 & 2 \\\hline
         GPUs & 2 & 8 & 8 \\\hline
         total train time (h) & 21 & 55 & 92 \\\hline
    \end{tabular}
    \caption{Summary of the hyperparameters used in the training of NVAE on the datasets used in this paper.}
    \label{tab:nvae-hyperparameters}
\end{table}
The figures of merit in Table~\ref{tab:nvae-hyperparameters} help
to understand the key novelty of NVAE, that is in the massive usage of space located latent variables. 
Consider for instance the case of Cifar10. 
The original input of dimension $32\times32\times3$ is first 
transformed to dimension $16\times 16\times 128$ and then, {\em without any further downscaling}, processed though a long sequence of residual cells
$(30\times2)$. At each iteration, 
a huge number of latent variables 
$(16\times16\times 20)$ is extracted and used for the internal
representation, which hence has a dimension widely 
larger than the input. Due to this fact, as it is also observed by the authors in the appendix, it is not surprising
that most of the variables will collapse during training.

Working with such a huge number of latent variables 
introduces a lot of issues; in particular, it becomes crucial
to balance the KL-component of variables belonging to different groups. 
To this aim, the authors introduce additional balancing coefficients $\gamma_l$ to ensure that a similar amount of
information is encoded in each group (see \cite{NVAE}, appendix A):
\[D_{KL}(q(z|x)||p(z)) = \sum_{l=1}^L \gamma_l \EX_{q(z<l|x)}
[D_{KL}(q(z_l|x, z_{<l})||p(z_l|z_{<l}))] \]
The balancing coefficient $\gamma_l$
is kept proportional to the KL term for that group, in such
a way to encourage the model to revive the latent variables in that group when KL is low, and to clip them if KL is too high.
Additionally, $\gamma_l$ is also proportional to the size of each group, to encourage the use of variables at lower scales.

NVAE architectures have a relatively small number of parameters, due to the extensive use of convolutions and depthwise separable convolutions; however, they require a massive amount of memory, 
and huge computational power: for the configuration used for Cifar10, composed by 30 groups at scale $16\time16$, we estimated a number of flops for the inference phase larger then 100G. 

Due to this reasons, we experimented a sensibly lighter architecture, just composed of 5 groups, with a few additional
convolutions to augment the receptive fileds of the spatially
located latent variables. The good news is that the network, 
even in this severely crippled form, still seems to learn; 
however, results are really modest and below the performances
of different networks with comparable complexity.

\subsubsection{HFVAE}
As we already remarked, the main novelty of NVAE is
in the massive exploitation of a huge number of spatially
located latent variables. In order to test the relevance 
of this architectural decision, we also tested a different
variant of the hereditary architecture of Figure \ref{fig:nvae}, where we drop the spatial dimension for latent variables, using instead a Featurewise Linear Modulation Layer \cite{Film} to modulate channels according to the internal representation. In addition, the first approximation $h_1$ is directly produced from the latent variable set $z_0$ through
a dense transformation. The general idea is that at lower scales we decide the content of the resulting image, while stylistic
details at different resolutions (usually captured in 
channels correlations \cite{style-transfer}) are added at 
higher scales.
We call this variant HFVAE (Hereditay Film VAE); a similar architecture
has been investigated in \cite{branca}.

\section{Numerical results}\label{sec:tables}
In this section, we provide quantitative evaluations for some configurations of the models previously discussed. The precise configurations (layers, channels, blocks, etc.) are discussed
below. 

The datasets used for the comparison are CIFAR10 and
CelebA: in a GreenAI perspective, we are reluctant to 
address more complex datasets, at higher resolutions,
that would require additional computational resources 
and additional costs. On CelebA, we just evaluated a 
subset of particularly interesting models.

For each model we provide the following figures:
\begin{description}
\item[{\bf params}:] the number of parameters;
\item[{\bf FLOPS}:] an estimation of number of FLOPS 
(see Section~\ref{sec:flops} for more details); 
\item[{\bf MSE}:] the mean reconstruction error $\times 10^3$;
\item[{\bf REC}:] the FID value computed over {\em reconstructed} images; 
\item[{\bf GEN1}:] the FID value computed over 
images generated by a first VAE;
\item[{\bf GEN2}:] the FID value computed taking advantage of a second VAE;
\item[{\bf GMM}:] the FID value computed by 
superimposing a GMM of ten Gaussians\footnote{Augmenting the number of Gaussians does not sensibly improve generation} on the latent space. In the case of hierarchical models, the GMM is computed on the innermost set of latent variables.
\end{description}

The following list provides a legenda for the names of models used in the following tables:
\begin{description}
\item[\bf{CNN-by-lz}] Vanilla VAE with CNN architecture, basedim of $y$ channels and latent space of dimension $z$.

\item[\bf{L2-RAE-by-lz}] $L_2$-RAE with CNN architecture, basedim of $y$ channels and latent space of dimension $z$.

\item[\bf{GP-RAE-by-lz}] GP-RAE with CNN architecture, basedim of $y$ channels and latent space of dimension $z$.

\item[\bf{Resnet-sx-by-lz}] Resnet-like model, with $x$ ScaleBlocks, a basedim of $y$ channels, and a latent space of dimension $z$.

\item[\bf{HFVAE-sx-by-lz}] HFVAE with x scales, ScaleBlocks, a basedim of y channels, and a latent space of dimension $z$ at hereditary scales; the base latent dimension $z_0$ is 64.

\item[\bf{NVAE-zx-by-lz}] NVAE with $x$ latent variables channels, a basedim of $y$ and $z$ latent groups of the same scale.
\end{description}

\begin{table}[ht]
\begin{center}
\begin{tabular}{|c|c|c|c|c|c|c|c|}\hline
model        & params & FLOPS & MSE & REC & GEN1 & GEN2 & GMM \\\hline
CNN-b128-l128 & 31,034,755 & 2,397M & 2.8 & 27.6 & 96.2 & 96.8 & 89.0 \\\hline
L2-RAE-b128-l128 & 30,510,339 & 2,395M & 1.2 & 9.9 & 108.1 & 88.4 & 78.2 \\\hline
GP-RAE-b128-l128 & 30,510,339 & 2,395M & 1.2 & 10.6 & 118.0 & 97.6 & 76.4 \\\hline
Resnet-s4-b48-l128 & 16,179,363 & 1,431M & 1.5 & 37.2 & 110.0 &  93.9 &  96.3 \\\hline
Resnet-s4-b48-l100 & 16,064,619 & 1,430M & 1.6 & 37.5 & 102.9 & 88.4 &  91.4 \\\hline
Resnet-s4-b64-l64 & 27,766,275 & 2,539M & 1.7 & 36.5 & 94.2 & 78.8  &  85.1 \\\hline
HFVAE-s4-z4-l48 & 27,139,755 & 1,163M & 1.8 & 45.9 & 93.3 & 90.8 & 90.0 \\\hline
HFVAE-s4-z12-l64 & 48,113,051 & 2,085M & 1.3 & 33.3 & 89.0 & 85.7 & 86.4 \\\hline
NVAE-z10-b100-l4 & 8,305,521 & 4,478M & 3.2 & 62.6 & 96.1 & 87.4 & 91.4 \\ \hline
\end{tabular}
\end{center}
\caption{Summary of the results obtained with the networks in the first column on Cifar10. The columns contains the the values of the parameters previously  described in the paper.}
\label{tab:Cifarresults}
\end{table}

\begin{table}[ht]
\begin{center}
\begin{tabular}{|c|c|c|c|c|c|c|c|}\hline
model        & params & FLOPS & MSE & REC & GEN1 & GEN2 & GMM \\\hline
CNN-b128-l64 & 40,668,419 & 4,104M & 3.2 & 48.4 & 66.9 & 56.2 & 55.2 \\\hline
L2-RAE-b128-l64 & 27,359,043 & 4,102M & 3.3 & 39.8 & 230.2 & 61.7 & 45.1 \\\hline
GP-RAE-b128-l64 & 27,359,043 & 4,102M & 3.2 & 41.2 & 230.6 & 65.3 & 47.0 \\\hline
Resnet-s4-b32-l64 & 19,330,627 & 2,924M & 2.8 & 51.4 & 66.0 & 54.9 & 57.4 \\\hline
Resnet-s4-b48-l64 & 38,996,003 & 6,452M & 2.5 & 46.8 & 61.7 & 50.8 & 54.5 \\\hline
Resnet-s3-b64-l64 & 21,370,179 & 5,949M & 2.6 & 39.2 & 59.3 & 44.9 & 45.8 \\\hline
\end{tabular}
\end{center}
\caption{Summary of the results obtained with the networks in the first column on CelebA. The columns contains the the values of the parameters previously  described in the paper.}
\label{tab:Celebaresults}
\end{table}

\subsection{Discussion}\label{sec:discussion}
Here we draw a few conclusions about the design of Variational Autoencoders deriving from the previous investigation (Tables \ref{tab:Cifarresults} and \ref{tab:Celebaresults}) and our past experience with VAEs.

\begin{itemize}
    \item The decoder is more important than the encoder. For instance, in
    the ResNet architecture latent features are extracted via a 
    GlobalAverage layer, obtaining robust features, 
    less prone to overfitting.
    \item Working with a larger number of latent variables improves reconstruction, but this does not eventually implies better generation.
    This is e.g. evident comparing the two Resnet-like architectures with
    latent spaces of dimension 128 and 100. 
    \item Fitting a GMM over the latent space \cite{deterministic} is a cheap technique (it just takes a few minutes) that invariably improves generation, both in terms 
    of perceptual quality and FID score. This fact also confirms the
    mismatch between the prior and the aggregated posterior discussed in
    Section~\ref{sec:mismatch}.
   \item The second stage technique \cite{TwoStage} typically requires 
   some tuning in order
   to properly works, but when it does it usually outperforms the GMM approach. Tuning may involve the loss function (we used cosine similarity
   in this work), the architecture of the second VAE, and the learning rate
   (more generally, the optimizer's parameters).
   \item Hierarchical architectures are complex systems, difficult 
   to understand and to work with (monitoring/calibrating training is 
   a really complex task). We cannot express an opinion about NVAE, since 
   its complexity
   trespasses our modest computational facilities, but simpler architectures
   like those described in \cite{DRAW} or \cite{Eslami18}, in our experience, 
   do not sensibly improve generation over a well constructed traditional VAE. 
   \item The loss of variance for generated images \cite{varianceloss} (see Section~\ref{sec:blurriness}) is confirmed in all models, and it almost coincides with the mean squared error for reconstruction.
   
\end{itemize}

\subsection{Energetic evaluation}
Before comparing the energetic footprint of the different models, let us briefly discuss the notion of FLOPS as a measure of 
computational efficiency. FLOPS have been computed by a library for Tensorflow Keras under development at the University of Bologna, and inspired by similar works for PyTorch (see e.g.
\url{https://github.com/sovrasov/flops-counter.pytorch}).
FLOPS only provide an abstract, machine independent notion of complexity; typically, only the most expensive layers are taken
into account (those with superlinear complexity with respect to the size of inputs). The way this quantity will result in an actual execution time and energetic consumption does however largely depend 
from the underlying hardware, and the available parallelism.
As an example, in Table~\ref{tab:time comparison} we compare the execution time for a 
forward step over the test set of Cifar10 (10K) for a couple 
of hardware configurations. The first one is a Laptop with an
NVIDIA Quadro T2000 graphics card and a cpu Core i7-9850H;
the second one is a workstation with an Asus GeForce 
DUAL-GTX1060-O6G graphic card and a cpu intel Core i7-7700K.
Observe the strong dependency from
the batchsize, that is not surprising but worth to be recalled
(see \cite{evaluationNN} for a critical analysis of the 
performance of Neural Networks architectures). Of course, as soon as
we move the computation on a cloud, execution times are practically unpredictable. 

\begin{table}[ht]
\begin{center}
\begin{tabular}{|c|c|c|c|c|c|}\hline
network & bs 100 & bs 10 & bs 1\\\hline
Resnet-s4-b48-l128 & $3.0 \pm .2 $ & $6.0 \pm .2$ & $33.3 \pm .4$\\
& $4.9\pm .2 $ & $8.8 \pm .2 $ & $49.5 \pm .5$\\\hline
Resnet-s4-b48-l100  & $2.86 \pm .1$ & $5.9 \pm .2$ & $32.9 \pm .4$ \\
& $4.8 \pm .2$ & $8.7 \pm .2$ & $49.1 \pm .5$ \\\hline
Resnet-s3-b64-l64 & $4.4 \pm .2$ & $9. \pm .3$ & $47.8 \pm .4$ \\
& $7.2 \pm .2$ & $13.5 \pm .2$ & $78.6 \pm .5$ \\\hline
\end{tabular}
\end{center}
\caption{\label{tab:time comparison}Average Forward Time over the Cifar10 TestSet (10K images) for different networks, hardware and batchsize. The two times entries in each cell refer to different
machines: the first is a Laptop with an
NVIDIA Quadro T2000 graphics card and a Core i7-9850H cpu;
the second is a workstation with an Asus GeForce 
DUAL-GTX1060-O6G graphic card and a intel Core i7-7700K cpu.}
\end{table}

Unfortunately, as we shall see, even for a {\em given} computational device, the relation between FLOPS and execution time is quite aleatory.

Following the traditional paradigm, we compare performances on the forward pass. This is already a questionable point; on one side, it is true that 
this reflects the final usage of the network when it is deployed in practical 
applications; on the other side, it is plausible to believe that training
still takes a prevalent part of the lifetime of any neural 
network. 
Restricting the investigation to forward time means not taking into account some expensive techniques of the training of modern systems, such as 
regularization components. For example, it is possible to notice that in Table \ref{tab:CifarTimes}, $L_2$-RAE and GP-RAE have exactly the same number of FLOPs, since in terms of forward execution they are equal. However, we highlight that  the training  of  GP-RAE is almost ten times slower than the training of  $L_2$-RAE. This is a consequence of the fact that the regularization term of GP-RAE involves the computation of the decoder gradient with respect to the latent variables, which is an expensive operation not required in $L_2$-RAE.  Consequently, even if the two models have more or less the same performance in terms of generation quality, $L_2$-RAE should be preferred, since its training is cheaper. Moreover, taking into account only the FLOPs of the model, the actual convergence speed of systems is neglected.

\begin{figure}[ht]
    \centering
    \includegraphics[width=.6\textwidth]{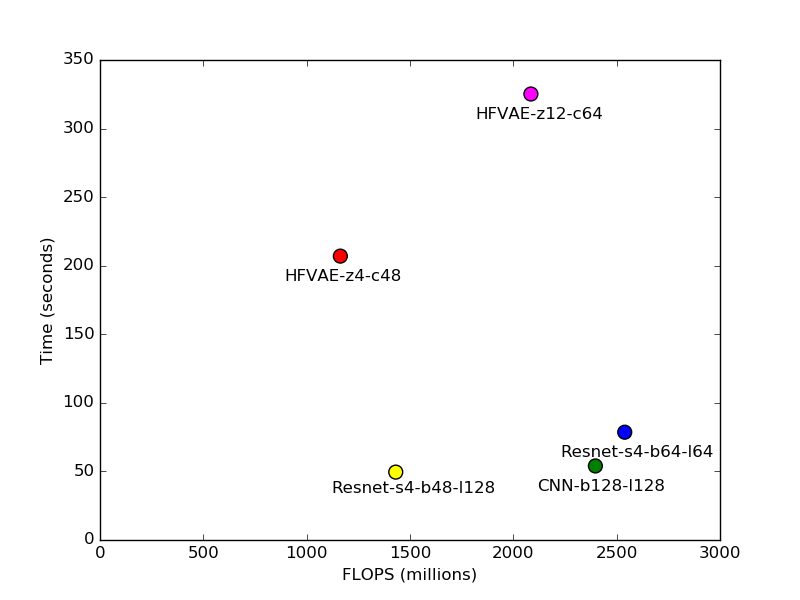}
    \caption{FLOPS versus Execution time. From the plot we can evince the little
    relation between the two figures but, possibly, at a magnitude level.}
    \label{fig:flops_vs_time}
\end{figure}

The results of the energetic evaluation on the forward pass
are given in Table~\ref{tab:CifarTimes}; inference times have been computed over a workstation  with  an  Asus  GeForceDUAL-GTX1060-O6G graphic card and a intel Core i7-7700K cpu. The same results have also been expressed in graphical form in Figure \ref{fig:flops_vs_time}, relatively to a batchsize of dimension 1. In the plot, we omit L2-RAE and GP-RAE, since their
architectures and figures are essentially analogous to the basic CNN; similarly for some Resnet architectures.

As it is clear from these results, there is no well defined correspondence between
FLOPS and execution time.  As observed by several authors (see e.g. \cite{trap_of_flops}), memory access time is another crucial factor in real implementations, as densely packed data might be read faster than a few numbers of scattered values. For instance, while depthwise convolutions greatly reduce the number of parameters and FLOPs, they require a more fragmented memory access, harder to be implemented efficiently.

\begin{table}[ht]
\begin{center}
\begin{tabular}{|c|c|c|c|c|c|}\hline
model        & params & FLOPS & time bs 100 & time bs 10 & time bs 1\\\hline
CNN-b128-l128 & 31,034,755 & 2,397M & 5.8 $\pm$ .1 & 9.0 $\pm$ .1 & 54.1 $\pm$ .4 \\\hline
L2-RAE-b128-l128 & 30,510,339 & 2,395M & 11.6 $\pm$ .2 & 13.9 $\pm$ .2 & 57.3 $\pm$ .5 \\\hline
GP-RAE-b128-l128 & 30,510,339 & 2,395M & 12.5 $\pm$ .2 & 14.1 $\pm$ .2 & 56.3 $\pm$ .5 \\\hline
Resnet-s4-b48-l128 & 16,179,363 & 1,431M & $4.9\pm .2 $ & $8.8 \pm .2 $ & $49.5 \pm .4$\\\hline
Resnet-s4-b48-l100 & 16,064,619 & 1,430M & $4.8 \pm .2$ & $8.7 \pm .2$ & $49.1 \pm .4$ \\\hline
Resnet-s4-b64-l64 & 27,766,275 & 2,539M & $7.2 \pm .2$ & $13.5 \pm .2$ & $78.6 \pm .5$ \\\hline
HFVAE-s4-z4-l48 & 27,139,755 & 1,163M & $13.1 \pm .2$  & $29.2 \pm .3 $ & $207.0 \pm 1.1$ \\\hline
HFVAE-s4-z12-l64 & 48,113,051 & 2,085M & $21.3 \pm .3 $ & $48.7 \pm .4$ & $325.2 \pm 1.6 $\\\hline
\end{tabular}
\end{center}
\caption{\label{tab:CifarTimes}Average Forward Time for 10000 inputs for Cifar10 architectures; times refer to a workstation equipped with an  Asus  GeForceDUAL-GTX1060-O6G graphic card and a intel Core i7-7700K cpu.}
\end{table} 

\section{Conclusions}\label{sec:conclusions}
In this article, we presented a critical survey of recent variants of Variational Autoencoders, referring them to 
the several problems that still hinder this generative paradigm. In view of the emerging GreenAI paradigm \cite{GreenAI}, we also focused the attention on the 
computational cost of the different architectures. 
The main conclusions of our investigation are given in
Section~\ref{sec:discussion}, and we shall not try to 
summarize them here; we just observe that, while we strongly support the GreenAI vision, we must eventually
find better metrics than FLOPS to compare the energetic performance 
of neural networks, or more realistic way to compute them.

The constant improvement in generative sampling during the last few years is very promising for the future of this field, suggesting that state-of-the-art generative
performance can be achieved or possibly even improved by
carefully designed VAE architectures. 

At the same time, the quest for scaling models to higher
resolution and larger images, and the introduction of additional, and usually computationally expensive, regularization techniques, is a scaring and dangerous perspective from the point of view of GreenAI. 

From this point of view, our experience with NVAE is explicative and quite frustrating. The architecture is interesting, and it should eventually deserve a deeper investigation; unfortunately, it seems 
to require computational facilities far beyond those at our 
disposal.\bigskip

\noindent
{\bf Acknowledgements} We would like to thank Federico Brunelllo who, under the supervision of Prof. Asperti, is developing the library for the computation of flops 
used in this article.\bigskip

\noindent
{\bf Conflict of Interest}
On behalf of all authors, the corresponding author states that there is no conflict of interest.

\bibliographystyle{plain}
\bibliography{variational}

\newpage
\appendix
\section{Examples of generated images}
\subsection{Cifar10}

\begin{figure}[h!]
\centering
\includegraphics[width=0.60\textwidth]{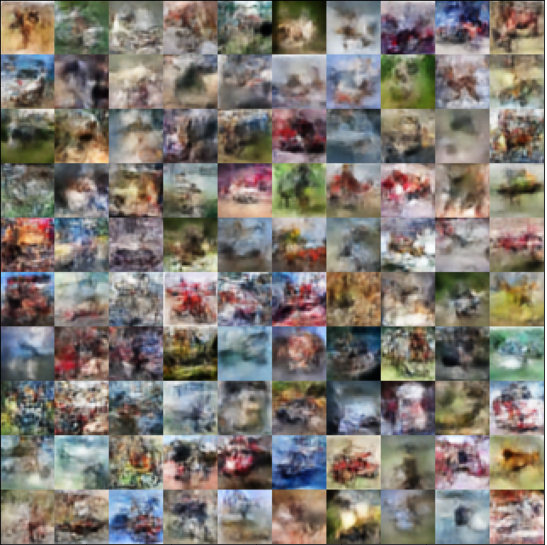}
\caption{Examples of Cifar-like images generated by Vanilla VAE.}
\end{figure}

\vspace{-30px}

\begin{figure}[h!]
\centering
\includegraphics[width=0.60\textwidth]{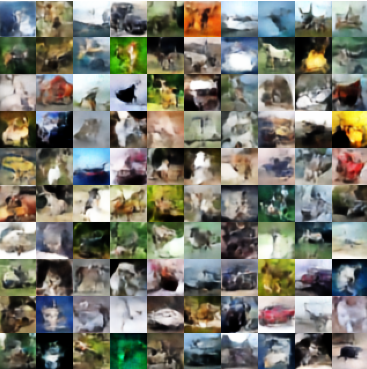}
\caption{Examples of Cifar-like images generated by Resnet.}
\end{figure}

\newpage

\begin{figure}[h!]
\centering
\includegraphics[width=0.65\textwidth]{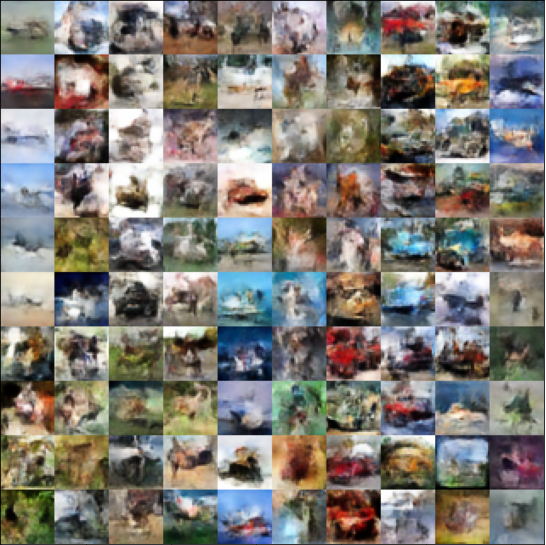}
\caption{Examples of Cifar-like images generated by $L_2$-RAE.}
\end{figure}

\begin{figure}[h!]
\centering
\includegraphics[width=0.65\textwidth]{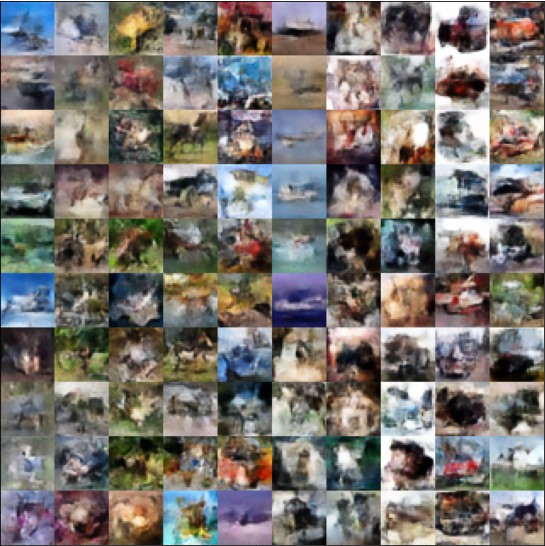}
\caption{Examples of Cifar-like images generated by GP-RAE.}
\end{figure}

\newpage

\begin{figure}[h!]
\centering
\includegraphics[width=0.65\textwidth]{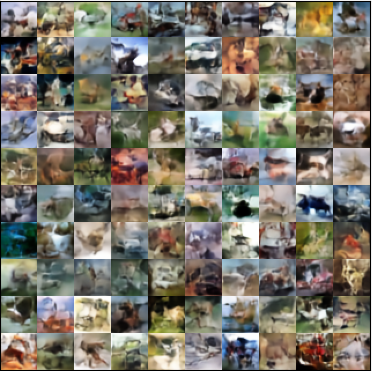}
\caption{Examples of Cifar-like images generated by HFVAE.}
\end{figure}

\begin{figure}[h!]
\centering
\includegraphics[width=0.65\textwidth]{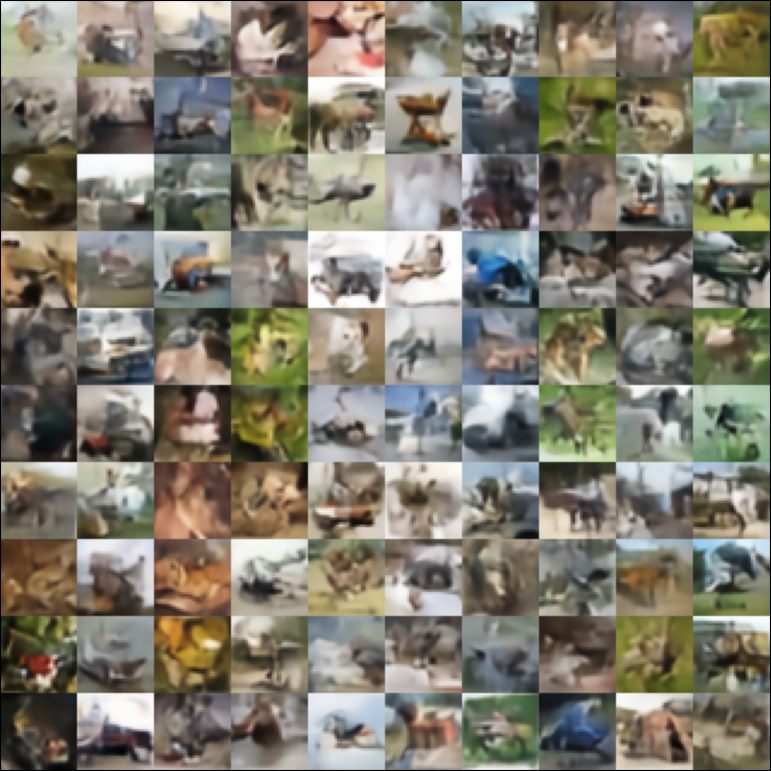}
\caption{Examples of Cifar-like images generated by NVAE.}
\end{figure}

\newpage

\subsection{CelebA}

\begin{figure}[h!]
\centering
\includegraphics[width=0.6\textwidth]{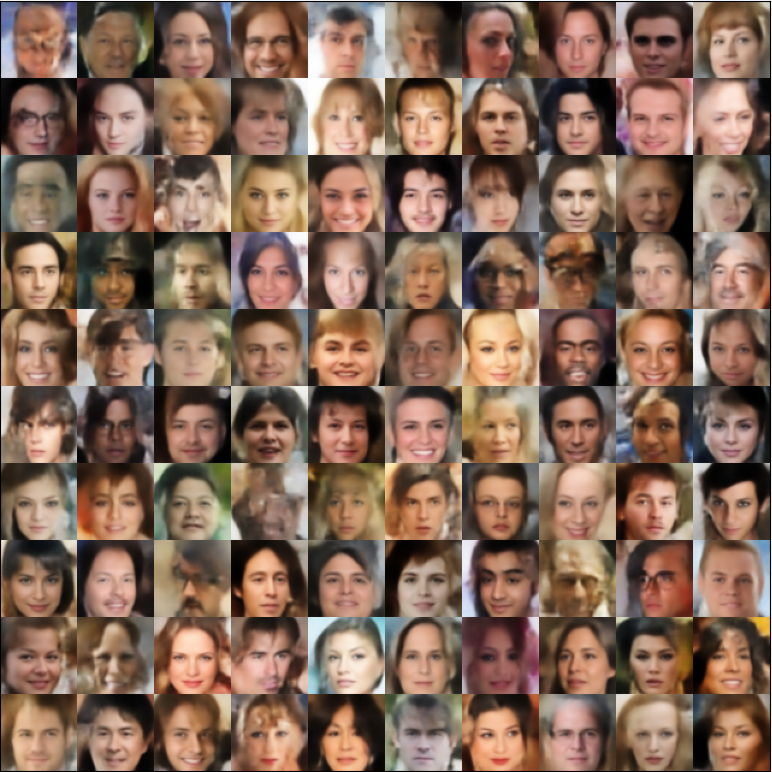}
\caption{Examples of CelebA faces generated by Vanilla VAE.}
\end{figure}

\begin{figure}[h!]
\centering
\includegraphics[width=0.85\textwidth]{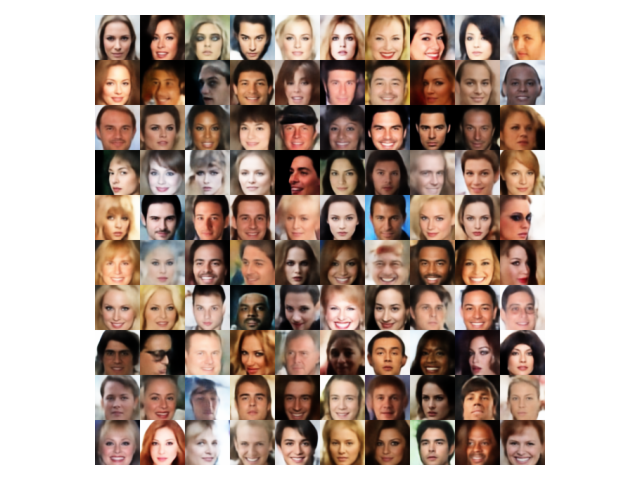}
\caption{Examples of CelebA faces generated by Resnet-s3-b64-l64.}
\end{figure}

\newpage

\begin{figure}[h!]
\centering
\includegraphics[width=0.65\textwidth]{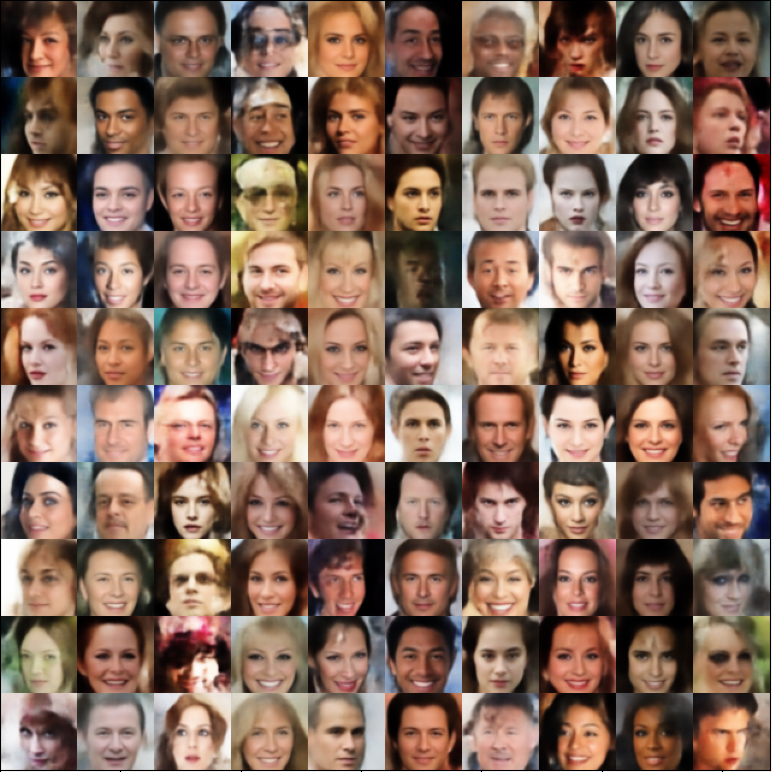}
\caption{Examples of CelebA faces generated by $L_2$-RAE.}
\end{figure}

\begin{figure}[h!]
\centering
\includegraphics[width=0.65\textwidth]{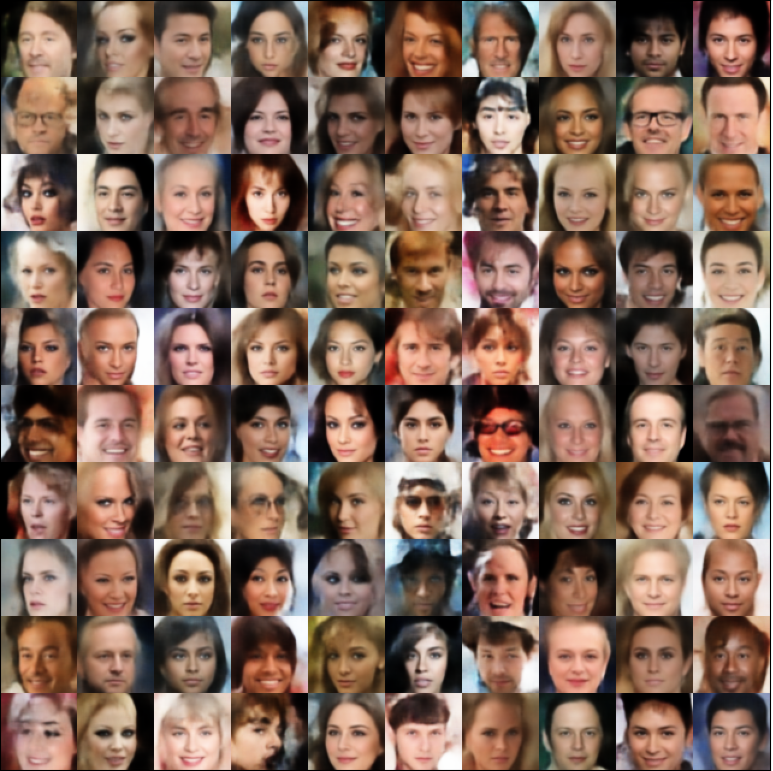}
\caption{Examples of CelebA faces generated by GP-RAE.}
\end{figure}

\end{document}